\documentclass[letterpaper]{article} 
\usepackage{aaai23}  
\usepackage{times}  
\usepackage{helvet}  
\usepackage{courier}  
\usepackage[hyphens]{url}  
\usepackage{graphicx} 
\usepackage{natbib}  
\usepackage{caption} 
\usepackage{bm}
\usepackage{multirow}
\usepackage{booktabs}
\usepackage{array}

\usepackage[labelformat=simple]{subcaption}

\usepackage{amstext}
\usepackage{amsmath}
\usepackage{makecell}
\usepackage{amsfonts}
\usepackage{float}

\usepackage[linesnumbered,ruled,vlined]{algorithm2e}
\SetKwInput{KwInput}{Input}                

\usepackage{newfloat}
\usepackage{listings}
\DeclareCaptionStyle{ruled}{labelfont=normalfont,labelsep=colon,strut=off} 
\floatstyle{ruled}
\newfloat{listing}{tb}{lst}{}
\floatname{listing}{Listing}
\title{Learning from Good Trajectories in Offline Multi-Agent Reinforcement Learning}
\author {
	Qi Tian\textsuperscript{\rm 1},
	Kun Kuang\textsuperscript{\rm 1,}\thanks{Corresponding author.},
	Furui Liu\textsuperscript{\rm 2},
	Baoxiang Wang\textsuperscript{\rm 3}
}
\affiliations {
	\textsuperscript{\rm 1} College of Computer Science and Technology, Zhejiang University, Hangzhou, China\\
	\textsuperscript{\rm 2} Huawei Noah’s Ark Lab, Beijing, China\\
	\textsuperscript{\rm 3} School of Data Science, Chinese University of Hong Kong (Shenzhen), Shenzhen, China\\
	\{tianqics,kunkuang\}@zju.edu.cn, liufurui2@huawei.com, bxiangwang@cuhk.edu.cn
}

\usepackage{bibentry}

\begin{document}

\maketitle

\begin{abstract}
	Offline multi-agent reinforcement learning (MARL) aims to learn effective multi-agent policies from pre-collected datasets, which is an important step toward the deployment of multi-agent systems in real-world applications.
	However, in practice, each individual behavior policy that generates multi-agent joint trajectories usually has a different level of how well it performs. \emph{e.g.}, an agent is a random policy while other agents are medium policies.
	In the cooperative game with global reward, one agent learned by existing offline MARL often inherits this random policy, jeopardizing the performance of the entire team.
	In this paper, we investigate offline MARL with explicit consideration on the diversity of agent-wise trajectories and propose a novel framework called Shared Individual Trajectories (SIT) to address this problem. 
	Specifically, an attention-based reward decomposition network assigns the credit to each agent through a differentiable key-value memory mechanism in an offline manner. 
	These decomposed credits are then used to reconstruct the joint offline datasets into prioritized experience replay with individual trajectories, thereafter agents can share their good trajectories and conservatively train their policies with a graph attention network (GAT) based critic.
	We evaluate our method in both discrete control (\emph{i.e.}, StarCraft II and multi-agent particle environment) and continuous control (\emph{i.e.}, multi-agent mujoco).
	The results indicate that our method achieves significantly better results in complex and mixed offline multi-agent datasets, especially when the difference of data quality between individual trajectories is large.
\end{abstract}

\section{Introduction}
Multi-agent reinforcement learning (MARL) has shown its powerful ability to solve many complex decision-making tasks. \emph{e.g.}, game playing \cite{samvelyan2019starcraft}. 
However, deploying MARL to practical applications is not easy since interaction with the real world is usually prohibitive, costly, or risky \cite{garcia2015comprehensive},
\emph{e.g.}, autonomous driving \cite{shalev2016safe}.
Thus offline MARL, which aims to learn multi-agent policies in the previously-collected, non-expert datasets without further interaction with environments, is an ideal way to cope with practical problems.

Recently, \citet{yang2021believe} first investigated offline MARL and found that multi-agent systems are more susceptible to extrapolation error \cite{fujimoto2019off}, \emph{i.e.}, a phenomenon in which unseen state-action pairs are erroneously estimated, compared to offline single-agent reinforcement learning (RL) \cite{fujimoto2019off,kumar2019stabilizing,wu2019behavior,kumar2020conservative,fujimoto2021minimalist}. Then they proposed implicit constraint Q-learning (ICQ), which can effectively alleviate this problem by only trusting the state-action pairs given in the dataset for value estimation.
\begin{figure}[t]
	\begin{center}
		\includegraphics[width=0.99\linewidth]{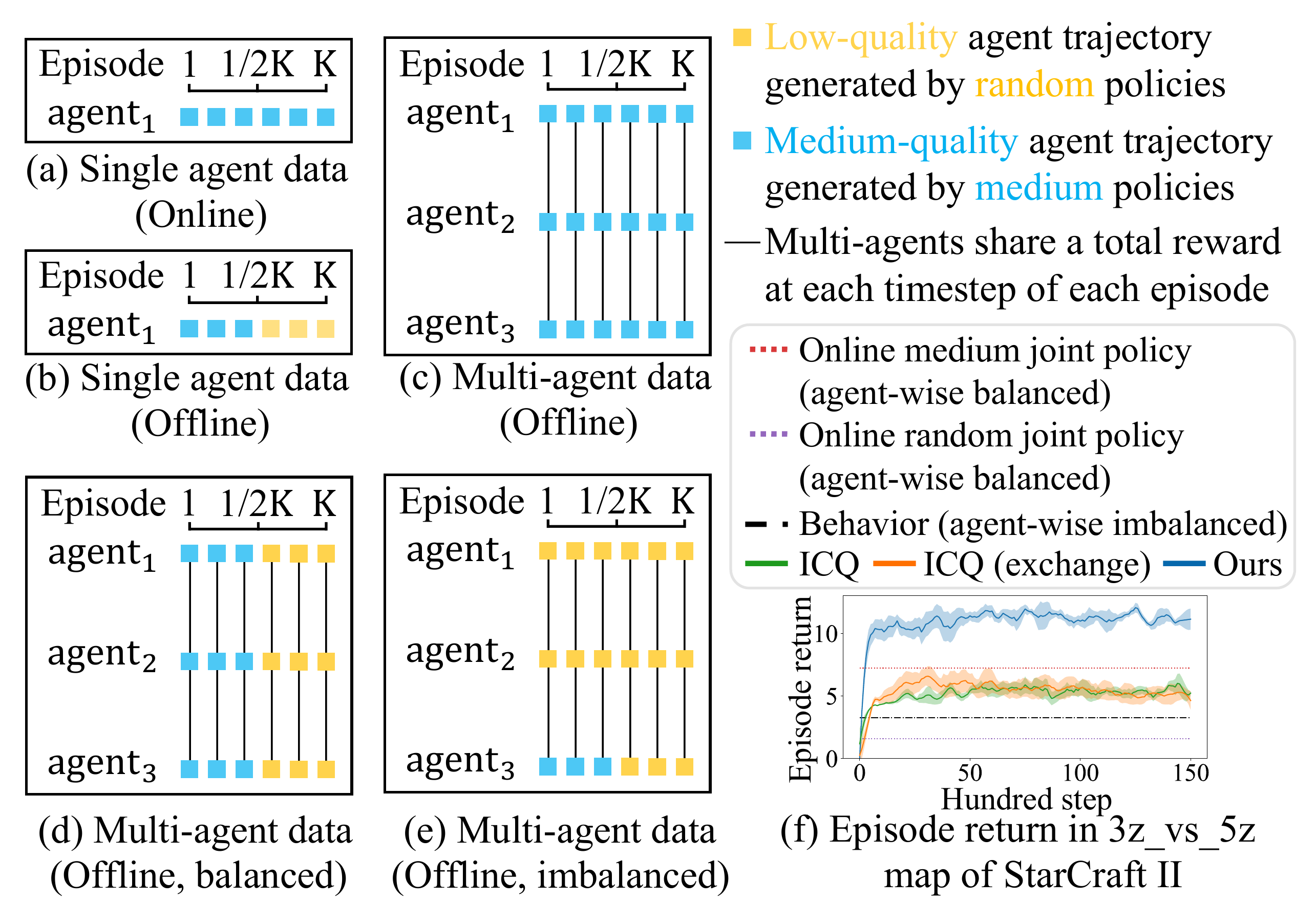}
		\caption{(a)$\sim$(e) Data composition of data replay in episodic online/offline RL/MARL, where the online learning agents in (a) and (c) are assumed to be medium policies. (f) Episode return in 3z\_vs\_5z map of StarCraft II with the offline data structure of (e).}
		\label{intro}
	\end{center}
\end{figure}

In this paper, we point out that in addition to extrapolation error, the diversity of trajectories in the offline multi-agent dataset also deserves attention since the composition of this dataset is much more complex than its single-agent version.
For better illustration, we briefly summarize the data quality of data replay in episodic online/offline RL/MARL as is shown in Figure \ref{intro}.
In online settings, since the data replay is updated rollingly based on the learning policy, the quality of all data points is approximately the same as is shown in Figure \ref{intro}(a) and Figure \ref{intro}(c), where line connections in Figure \ref{intro}(c) represent multi-agent systems only provide one global reward at each timestep of each episode.
In offline settings, there is no restriction on the quality of collected data, thus the single-agent data replay usually contains multi-source and suboptimal data as is shown in Figure \ref{intro}(b).
Figure \ref{intro}(d) directly extends the offline single-agent data composition to a multi-agent version.
However, in practical tasks, each individual trajectory in an offline multi-agent joint trajectory usually has a different data quality, as is shown in Figure \ref{intro}(e).
For example, consider the task that two robotic arms (agents) work together to lift a large circular object, and the reward is the height of the center of mass.
Suppose that two workers operate two robotic arms simultaneously to collect offline data, but they have different levels of proficiency in robotic arm operation. The offline joint data generated in this case is an agent-wise imbalanced multi-agent dataset.

To investigate the performance of the current state-of-the-art offline MARL algorithm, ICQ, on the imbalanced dataset, we test it on the 3s\_vs\_5z map in StarCraft II \cite{samvelyan2019starcraft}.
Specifically, as is shown in Figure \ref{intro}(f), we first obtain random joint policy (violet dotted) and medium joint policy (red dotted) through online training, and then utilize these two joint policies to construct an agent-wise imbalanced dataset with the data structure of Figure \ref{intro}(e).
We can observe that the performance of ICQ (green) is lower than that of the online medium joint policy (red dotted), as ${\rm agent_1}$ and ${\rm agent_2}$ may inherit the random behavior policies that generate the data.
One might attempt to solve this issue of ICQ by randomly exchanging local observations and actions among agents, so that ${\rm agent_1}$ and ${\rm agent_2}$ can share some medium-quality individual trajectories of ${\rm agent_3}$. Unfortunately, the performance of this alternative (orange) is similar to ICQ, since the proportion of the medium-quality individual data in the data replay does not change under the constraint of the total reward.

In this paper, we propose a novel algorithmic framework called Shared Individual Trajectories (SIT) to address this problem. 
It first learns an Attention-based Reward Decomposition Network with Ensemble Mechanism (ARDNEM), which assigns credits to each agent through a differentiable key-value memory mechanism in an offline manner.
Then these credits are used to reconstruct the original joint trajectories into Decomposed Prioritized Experience Replay (DPER) with individual trajectories, thereafter agents can share their good trajectories and conservatively train their policies with a graph attention network based critic.
Extensive experiments on both discrete control (\emph{i.e.}, StarCraft II and multi-agent particle environment) and continuous control (\emph{i.e.}, multi-agent mujoco) indicate that our method achieves significantly better results in complex and mixed offline multi-agent datasets, especially when the difference of data quality between individual trajectories is large.

\section{Related Work}
\label{Related Work}

\paragraph{Offline MARL.}
Recently, \citet{yang2021believe} first explored offline MARL and found that multi-agent systems are more susceptible to extrapolation error \cite{fujimoto2019off} compared to offline single-agent RL \cite{fujimoto2019off,kumar2020conservative,fujimoto2021minimalist}, and then they proposed implicit constraint Q-learning (ICQ) to solve this problem.
\citet{jiang2021offline} investigated the mismatch problem of transition probabilities in fully decentralized offline MARL.
However, they assume that each agent makes the decision based on the global state and the reward output by the environment can accurately evaluate each agent's action.
This is fundamentally different from our work since we focus on the partially observable setting in global reward games \cite{chang2003all}, which is a more practical situation.
Other works \cite{meng2021offline,jiang2021online} have made some progress in offline MARL training with online fine-tuning. 
Unfortunately, all the existing methods neglect to investigate the diversity of trajectories in offline multi-agent datasets, while we take the first step to fill this gap.

\paragraph{Multi-agent credit assignment.} 
In cooperative multi-agent environments, all agents receive one total reward.
The multi-agent credit assignment aims to correctly allocate the reward signal to each individual agent for a better groups’ coordination \cite{chang2003all}.
One popular class of solutions is value decomposition, which can decompose team value function into agent-wise value functions in an online fashion under the framework of the Bellman equation \cite{sunehag2018value,rashid2018qmix,yang2020qatten,li2021shapley}.
Different from these works, in this paper, we focus on explicitly decomposing the total reward into individual rewards in an offline fashion under the regression framework, and these decomposed rewards will be used to reconstruct the offline prioritized dataset.

\paragraph{Experience replay in RL/MARL.}
Experience replay, a mechanism for reusing historical data \cite{lin1992self}, is widely used in online RL \cite{wang2020deep}. 
Many prior works related to it focus on improving data utilization \cite{schaul2016prioritized,zha2019experience,oh2020learning}.
\emph{e.g.}, prioritized experience replay (PER) \cite{schaul2016prioritized} takes temporal-difference (TD) error as a metric for evaluating the value of data and performs importance sampling according to it.
Unfortunately, this metric fails in offline training due to severe overestimation in offline scenarios.
In online MARL, most works related to the experience replay focus on stable decentralized multi-agent training \cite{foerster2017stabilising,omidshafiei2017deep,palmer2018lenient}, 
but these methods usually rely on some auxiliary information, \emph{e.g.}, training iteration number, timestamp and exploration rate, which often are not provided by the offline settings.
SEAC \cite{christianos2020shared} proposed by Christianos et al. is most related to our work as it also shares individual trajectories among agents during online training.
However, SEAC assumes that each agent can directly obtain a local reward and does not consider the importance of each individual trajectory.
Instead, we need to decompose the global reward into local rewards in an offline manner, and then determine the quality of the individual trajectory based on the decomposed rewards for priority sampling.

\section{Preliminaries}
\label{Preliminaries and Notation}

\begin{figure*}[t]
	\begin{center}
		\includegraphics[width=\linewidth]{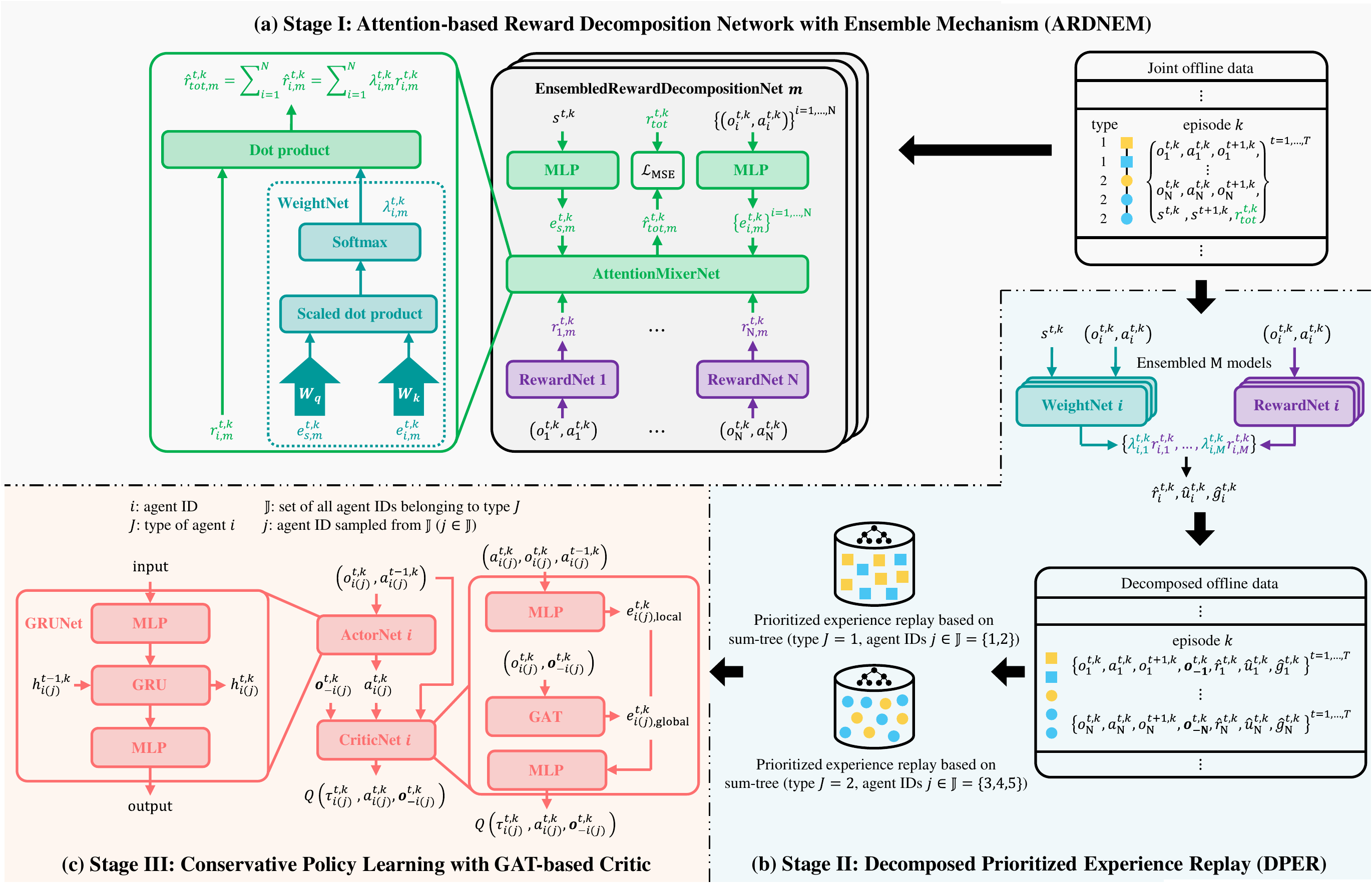}
		\caption{Overall framework of Shared Individual Trajectories (SIT) in offline MARL}
		\label{method}
	\end{center}
\end{figure*}

A fully cooperative multi-agent task in the global reward game \cite{chang2003all} can be described as a Decentralized Partially Observable Markov Decision Process (Dec-POMDP) \cite{oliehoek2016concise} consisting of a tuple $\langle \emph{N}, \mathcal{S}, \mathcal{A}, \mathcal{T}, \mathcal{R}, \mathcal{O}, \Omega, \gamma \rangle$.
Let $\emph{N}$ represent the number of agents and
$\mathcal{S}$ represent the true state space of the environment.
At each timestep $t \in \mathcal{Z}^{+}$ of episode $k \in \mathcal{Z}^{+}$, each agent $i \in \emph{N} \equiv \left\{ 1,\dots, n \right\}$ takes an action $a_i \in \mathcal{A}$, forming a joint
action $\boldsymbol{a} \in \bm{\mathcal{A}} \equiv \mathcal{A}^n$.
Let $\mathcal{T}(s'|s, \boldsymbol{a}): \mathcal{S} \times \bm{\mathcal{A}} \rightarrow \mathcal{S}$ represent the state transition function.
All agents share the global reward function $r(s,\boldsymbol{a}):\mathcal{S} \times \bm{\mathcal{A}} \rightarrow \mathcal{R}$ and $\gamma \in [ 0,1 )$ represents the discount factor.
Let $J$ represent the type of agent $i$, which is the prior knowledge given by the environment \cite{wang2019action,yang2020q}.
All agent IDs belonging to type $J$ are denoted as $j \in \mathbb{J}$, where $\mathbb{J}$ represents the set of IDs of all agents isomorphic to agent $i$ (including $i$).
We consider a partially observable setting in which each agent receives an individual observation $o_i \in \Omega$ according to the observation function $\mathcal{O}(s,a):\mathcal{S} \times \mathcal{A} \rightarrow \Omega$.
Each agent has an action-observation history $\tau_i \in \bm{T} \equiv \left(
\Omega \times \mathcal{A} \right)^{t} $, on which it conditions a stochastic policy $\pi_i \left( a_i|\tau_i \right)$. 
Following \citet{gronauer2022multi}, we focus on episodic training in this paper.

\section{Methodology}
\label{Method}

In this section, we propose a new algorithmic framework called Shared Individual Trajectories (SIT) in offline MARL, which can maximally exploit the useful knowledge in the complex and mixed offline multi-agent datasets to boost the performance of multi-agent systems.
SIT consists of three stages as shown in Figure \ref{method}:
Stage I: learning an Attention-based Reward Decomposition Network with Ensemble Mechanism (ARDNEM).
Stage II: reconstructing the original joint offline dataset into Decomposed Prioritized Experience Replay (DPER) based on the learned ARDNEM.
Stage III: conservative offline actor training with a graph attention network (GAT) based critic on DPER.

\subsection{Attention-Based Reward Decomposition Network with Ensemble Mechanism}

It is a non-trivial problem to evaluate the behavior of individual trajectories in an offline dataset under the global reward games since only one total reward can be accessed in this case.
We propose a reward decomposition network, as is shown in Figure \ref{method}(a), to solve this problem. Specifically, the individual reward $r_{i}^{t,k}$ at timestep $t$ of episode $k$ can be approximately estimated from the local observation $o_{i}^{t,k}$ and action $a_{i}^{t,k}$ through the reward network $f_i$. \emph{i.e.}, $r_{i}^{t,k}=f_i(o_{i}^{t,k},a_{i}^{t,k})$. 
In order to learn the contribution weight $\lambda_{i}^{t,k}$ of each individual reward $r_{i}^{t,k}$ to the total reward, we use the attention mechanism (\emph{i.e.}, the differentiable key-value memory model \cite{oh2016control}) to achieve this goal.
That is, we first encode the global state $s^{t,k}$ and local observation-action pair $(o^{t,k}_{i},a^{t,k}_{i})$ into embedding vectors $e^{t,k}_s$ and $e^{t,k}_i$ with two multi-layer perceptrons (MLPs) respectively, then we pass the similarity value between the global state's embedding vector $e^{t,k}_s$ and the individual features' embedding vector $e^{t,k}_i$ into a softmax
\begin{equation}
	\lambda_{i}^{t,k} \propto \exp ((e_s^{t,k})^T W_q^T W_k e_i^{t,k})\,,
\end{equation}
where the learnable weight $W_q$ and $W_k$ transform $e_s^{t,k}$ and $e_i^{t,k}$ into the query and key.
Thus the estimated total reward $\hat{r}_{tot}^{t,k}$ at each timestep $t$ of episode $k$ can be expressed as
\begin{equation}
	\hat{r}_{tot}^{t,k}=\sum_{i=1}^{N} \lambda_{i}^{t,k} r_{i}^{t,k}=\sum_{i=1}^{N} \lambda_{i}^{t,k} \cdot f_i(o_{i}^{t,k},a_{i}^{t,k})\,.
\end{equation}

Since misestimated individual rewards will have a large impact on agent learning, we want to obtain the uncertainty corresponding to each estimated value for correcting subsequent training.
Following \citet{chua2018deep}, we introduce ensemble mechanism \cite{osband2016deep} to meet this goal.
\emph{i.e.},  $M$ decomposition networks are learned simultaneously.
Finally, our Attention-based Reward Decomposition Network with Ensemble Mechanism (ARDNEM) w.r.t. parameters $\psi$ can be trained on the original joint offline dataset with the following mean-squared error (MSE) loss
\begin{equation}
	\mathcal{L}_{{\rm MSE}}(\psi)=\frac{1}{M} \sum_{m=1}^{M} \left(\sum_{i=1}^{N} \lambda_{i,m}^{t,k} \cdot f_{i,m}(o_{i}^{t,k},a_{i}^{t,k})-r_{tot}^{t,k}\right)^2\,,
\end{equation}
where $r_{tot}^{t,k}$ represents the true total reward in the offline dataset. In practice, the reward network parameters of different agents are shared for the scalability of our method.

\subsection{Decomposed Prioritized Experience Replay}
As is shown in Figure \ref{method}(b), after ARDNEM is learned, we can use it to estimate individual rewards for all local trajectories.
Specifically, since ARDNEM adopts the ensemble mechanism, we take the mean of $M$ model output $\lambda_{i,m}^{t,k} r_{i,m}^{t,k}$ as the estimation of the weighted individual reward $\hat{r}_{i}^{t,k}$:
\begin{equation}
	\hat{r}_{i}^{t,k}=\frac{1}{M} \sum_{m=1}^{M} \lambda_{i,m}^{t,k} r_{i,m}^{t,k} \,.
\end{equation}
Its corresponding variance is defined as the uncertainty $\hat{u}_i^{t,k}$ of the model prediction:
\begin{equation}
	\hat{u}_i^{t,k}=\sqrt{\frac{1}{M} \sum_{m=1}^{M} \left( \lambda_{i,m}^{t,k} r_{i,m}^{t,k} - \hat{r}_{i}^{t,k} \right)^2} \,.
\end{equation}

To maximally exploit the useful knowledge in the data replay, we need a metric to distinguish the importance of each data.
Many previous works in online RL do this through temporal-difference (TD) error \cite{schaul2016prioritized,zha2019experience,oh2020learning}. However, the value function suffers from severe overestimation in offline settings \cite{fujimoto2019off}, thus TD error is not an ideal choice. 
In this paper, considering that offline datasets are static and limited, we believe that high-quality data should be valued.
Therefore, we use Monte Carlo return $\hat{g}_i^{t,k}$ to measure the importance (or quality) of the data at each timestep $t$ of episode $k$:
\begin{equation}
	\hat{g}_i^{t,k}=\sum_{t=t^{\prime}}^{T} \gamma^{t-t^{\prime}} \hat{r}_{i}^{t,k}\,,
\end{equation}
where $\gamma$ represents discounted factor.

After the above efforts, each episode $k$ in the original joint trajectories (\emph{i.e.}, $\{o_{1}^{t,k},a_{1}^{t,k},o_{1}^{t+1,k},\dots,o_{{\rm N}}^{t,k},a_{{\rm N}}^{t,k},o_{{\rm N}}^{t+1,k},$
$s^{t,k},s^{t+1,k},r_{tot}^{t,k} \}^{t=1,\dots,T}$) can be decomposed to the agent-wise individual trajectories (\emph{i.e.}, $\{ \{o_{i}^{t,k},a_{i}^{t,k},o_{i}^{t+1,k},\bm{o_{-i}}^{t,k},$
$\hat{r}_{i}^{t,k},\hat{u}_i^{t,k},\hat{g}_i^{t,k} \}^{t=1,\dots,T}\}^{i=1,\dots,{\rm N}}$), where $\bm{o_{-i}}^{t,k}$ represents all local observations except for agent $i$.
We then store all these individual trajectories into single/multiple Decomposed Prioritized Experience Replay (DPER) according to their agent type.
Since we train multi-agent policies in an episodic manner, the mean of Monte Carlo return $\hat{g}_i^{t,k}$ for each episode $k$ is used as the sampling priority $\hat{p}_i^{k}=\frac{1}{T} \sum_{t=1}^{T} \hat{g}_i^{t,k}$.
To trade off priority sampling and uniform sampling, we use a softmax function with a temperature factor $\alpha$ to reshape the priorities in the decomposed dataset:
\begin{equation} \label{eq_alpha}
	p_i^{k}=\frac{e^{\hat{p}_i^{k}/\alpha}}{\sum_{j\in \mathbb{J},k}^{K} e^{\hat{p}_{j}^{k}/\alpha}} \,,
\end{equation}
where $\mathbb{J}$ represents the set of IDs of all agents isomorphic to agent $i$ (including $i$).
Temperature factor $\alpha$ determines how much prioritization is used, with $\alpha \rightarrow \infty$ corresponding to the uniform sampling.

In practice, considering the large gap in rewards for different environments,
all sampling priority $\hat{p}_i^{k}$ are linearly scaled to a uniform range, allowing our method to share the same temperature factor $\alpha$ across environments.
Meanwhile, we use the sum-tree \cite{schaul2016prioritized} as the storage structure of DPER to improve the sampling efficiency.
\begin{table*}[t]
	\setlength\tabcolsep{6pt}
	\small
	\centering
	\begin{tabular}{ccccccccc}
		\hline
		\multicolumn{9}{c}{StarCraft II (SC II)}                                                                                                                     \\ \hline
		&                 & Behavior & BC            & QMIX          & MABCQ         & MACQL         & ICQ           & Ours          \\ \hline \specialrule{0em}{1pt}{1pt}
		\multirow{6}{*}{\vspace{-1pt}\shortstack{Low\\Quality}} & 2s\_vs\_1sc     & 2.8     & 3.2$\pm$1.2     & 1.1$\pm$0.0     & 1.7$\pm$0.0     & 4.5$\pm$0.1     & 4.4$\pm$0.1     & \textbf{10.5$\bm{\pm}$0.4}    \\
		& 3s\_vs\_5z      & 2.9     & 2.9$\pm$0.4     & 2.3$\pm$0.2     & 4.5$\pm$0.2     & 4.5$\pm$0.1     & 5.2$\pm$0.1     & \textbf{11.1$\bm{\pm}$0.8}    \\
		& 2s3z            & 3.2     & 3.1$\pm$1.7     & 2.6$\pm$0.0     & 4.7$\pm$0.7     & 5.7$\pm$0.4     & 9.1$\pm$0.3     & \textbf{12.3$\bm{\pm}$1.6}    \\
		& 8m              & 3.1     & 3.1$\pm$0.2     & 2.0$\pm$0.6     & 5.7$\pm$1.2     & 2.5$\pm$0.6     & 5.5$\pm$0.5     & \textbf{10.8$\bm{\pm}$0.4}    \\
		& 1c3s5z          & 5.5     & 5.9$\pm$0.8     & 2.4$\pm$1.9     & 8.4$\pm$1.2     & 6.1$\pm$1.2     & 9.4$\pm$0.0     & \textbf{13.7$\bm{\pm}$0.3}    \\
		& 10m\_vs\_11m    & 3.8     & 4.2$\pm$0.7     & 0.7$\pm$0.5     & 4.3$\pm$1.8     & 5.2$\pm$0.1     & 8.0$\pm$0.2     & \textbf{12.3$\bm{\pm}$0.7}    \\ \specialrule{0em}{1pt}{1pt} \hline \specialrule{0em}{1pt}{1pt}
		\multirow{6}{*}{\shortstack{Medium\\Quality}} & 2s\_vs\_1sc     & 9.8     & 9.8$\pm$0.2     & 3.6$\pm$1.6     & 6.6$\pm$1.6     & 9.8$\pm$0.1     & 10.2$\pm$0.0    & \textbf{16.5$\bm{\pm}$1.8}    \\
		& 3s\_vs\_5z      & 7.0     & 7.6$\pm$0.4     & 2.6$\pm$1.6     & 5.8$\pm$0.8     & 8.8$\pm$1.4     & 13.5$\pm$1.2    & \textbf{17.8$\bm{\pm}$1.5}    \\
		& 2s3z            & 7.0     & 7.1$\pm$0.3     & 2.3$\pm$0.5     & 5.8$\pm$1.2     & 7.1$\pm$0.5     & 9.1$\pm$0.3     & \textbf{14.8$\bm{\pm}$0.3}    \\
		& 8m              & 9.8     & 9.6$\pm$0.3     & 2.4$\pm$0.0     & 4.0$\pm$1.6     & 10.5$\pm$1.2    & 13.4$\pm$0.6    & \textbf{14.8$\bm{\pm}$1.1}    \\
		& 1c3s5z          & 10.3    & 10.0$\pm$0.1    & 8.1$\pm$1.6     & 11.8$\pm$2.0    & 10.1$\pm$0.2    & 12.7$\pm$0.3    & \textbf{15.4$\bm{\pm}$0.4}    \\
		& 10m\_vs\_11m    & 9.2     & 9.2$\pm$0.3     & 1.7$\pm$0.4     & 3.5$\pm$0.6     & 9.5$\pm$0.5     & 10.6$\pm$0.4    & \textbf{13.6$\bm{\pm}$1.3}    \\ \hline
		\multicolumn{9}{c}{Multi-agent Particle Environment (MPE)}                                                                                                   \\ \hline
		&                 & Behavior & BC            & QMIX          & MABCQ         & MACQL         & ICQ           & Ours          \\ \hline \specialrule{0em}{1pt}{1pt}
		\multirow{3}{*}{\vspace{-2pt}\shortstack{Low\\Quality}} & CN\_3ls3l    & -157.7  & 161.1$\pm$4.2   & -231.9$\pm$3.0  & -174.3$\pm$22.7 & -217.2$\pm$18.1 & -117.8$\pm$11.5 & \textbf{-92.4$\bm{\pm}$7.2}   \\
		& CN\_4ls4l    & -278.4  & -274.7$\pm$6.0  & -316.4$\pm$17.3 & -194.5$\pm$8.3  & -300.5$\pm$7.3  & -231.4$\pm$2.2  & \textbf{-120.7$\bm{\pm}$13.3} \\
		& PP\_3p1p    & -249.5  & -253.7$\pm$10.8 & -336.9$\pm$22.9 & -239.9$\pm$32.3 & -326.5$\pm$18.6 & -227.7$\pm$6.7  & \textbf{-105.6$\bm{\pm}$9.0}  \\ \specialrule{0em}{1pt}{1pt} \hline \specialrule{0em}{1pt}{1pt}
		\multirow{3}{*}{\shortstack{Medium\\Quality}} & CN\_3ls3l    & -107.4  & -111.0$\pm$1.8  & -256.1$\pm$3.9  & -107.3$\pm$13.9 & -231.9$\pm$13.2 & -83.0$\pm$3.2   & \textbf{-54.5$\bm{\pm}$2.9}   \\
		& CN\_4ls4l    & -166.2  & -161.4$\pm$5.4  & -290.0$\pm$5.3  & -146.0$\pm$8.3  & -282.8$\pm$14.0 & -184.6$\pm$10.2 & \textbf{-72.3$\bm{\pm}$2.3}   \\
		& PP\_3p1p    & -155.4  & -156.2$\pm$7.3  & -296.6$\pm$28.8 & -230.2$\pm$27.9 & -272.5$\pm$14.4 & -158.6$\pm$5.2  & \textbf{-80.4$\bm{\pm}$4.7}   \\ \hline
		\multicolumn{9}{c}{Multi-Agent mujoco (MAmujoco)}                                                                                                            \\ \hline
		&                 & Behavior & BC            & FacMAC        & MABCQ         & MACQL         & ICQ           & Ours          \\ \hline \specialrule{0em}{1pt}{1pt}
		\multirow{2}{*}{\vspace{-3pt}\shortstack{Low\\Quality}} & HalfCheetah\_2l & -110.5  & -110.3$\pm$1.1  & -152.5$\pm$18.5 & -100.9$\pm$2.8  & -70.8$\pm$29.1  & -109.3$\pm$3.1  & \textbf{-0.3$\bm{\pm}$0.1}    \\
		& Walker\_2l      & -21.6   & -27.9$\pm$12.4  & -34.3$\pm$10.3  & -28.7$\pm$14.0  & 16.8$\pm$39.1   & -21.2$\pm$8.5   & \textbf{105.8$\bm{\pm}$20.8}  \\ \specialrule{0em}{1pt}{1pt} \hline \specialrule{0em}{1pt}{1pt}
		\multirow{2}{*}{\vspace{-1pt}\shortstack{Medium\\Quality}} & HalfCheetah\_2l & 41.7    & 45.3$\pm$5.4    & -95.3$\pm$45.6  & 64.9$\pm$15.0   & 20.3$\pm$34.7   & 50.4$\pm$18.9   & \textbf{164.1$\bm{\pm}$13.0}  \\
		& Walker\_2l      & 71.6    & 80.3$\pm$19.9   & -11.7$\pm$5.7   & 87.0$\pm$17.1   & 41.0$\pm$21.9   & 75.8$\pm$12.5   & \textbf{167.1$\bm{\pm}$36.6}  \\ \hline
	\end{tabular}
	\caption{The mean and variance of the episodic return on agent-wise imbalanced multi-agent datasets of various maps.}
	\label{result}
\end{table*}

\subsection{Conservative Policy Learning with GAT-Based Critic}
In this subsection, we will use the obtained type-wise DPER for multi-agent policy learning under the centralized training decentralized execution (CTDE) paradigm.
As is shown in Figure \ref{method}(c),
the input of the centralized critic for each agent $i$ consists of local information and global information.
Specifically, the former includes the local action-observation history $\tau^{t,k}_{i}=(o^{t,k}_{i},a^{t-1,k}_{i})$ \cite{peng2021facmac} and current action $a_i^{t,k}$ of each agent $i$.
We encode them into the local embedding vector $e_{i,{\rm local}}^{t,k}$ with two MLPs $f_{{\rm local}}$. \emph{i.e.}, $e_{i,{\rm local}}^{t,k}=f_{{\rm local}}(\tau^{t,k}_{i},a_i^{t,k})$.
The latter includes local observations of all agents $(o_i^{t,k},\bm{o_{-i}}^{t,k})$.
We construct these observations as a fully connected graph and aggregate the global embedding vector $e_{i,{\rm global}}^{t,k}$ via a graph attention network (GAT) \cite{velivckovic2018graph}, as
\begin{equation}
	\begin{aligned}
		& w_{i,j}^{t,k} = \frac{\exp(\text{LeakyReLU}(W_{2}^{T}[W_{1}o_{i}^{t,k};W_{1}o_{j}^{t,k}]))}{\sum_{k \in N}\exp(\text{LeakyReLU}(W_{2}^{T}[W_{1}o_{i}^{t,k};W_{1}o_{k}^{t,k}]))} \\
		& e_{i,{\rm global}}^{t,k} = {\textstyle \sum_{j \in N} w_{i,j}^{t,k}W_{1}o_{j}^{t,k}} \,,
	\end{aligned}
\end{equation}
where $W_1$ and $W_2$ represent the learnable weights in GAT. $(\cdot)^{T}$ represents transposition. $[\cdot;\cdot]$ represents concatenation operation.
Then, the centralized critic of each agent $i$ can be expressed as $Q_i(\tau_{i}^{t,k},a_{i}^{t,k},\bm{o_{-i}}^{t,k})=f_{{\rm agg}}([e_{i,{\rm local}}^{t,k};e_{i,{\rm global}}^{t,k}])$, where $f_{{\rm agg}}$ represents the aggregation network with two MLPs. In order to simplify the expression, we denote the critic of agent $i$ as $Q_i$ in our subsequent description.

To alleviate the severe extrapolation error in offline agent learning, we plug the filtering mechanism of CRR \cite{wang2020critic} into individual policy learning.
This method can implicitly constrain the forward KL divergence between the learning policy and the behavior policy, which is widely used in offline single-agent learning \cite{wang2020critic,nair2020awac,gulcehre2021regularized} and multi-agent learning \cite{yang2021believe}.
Formally, suppose that the type of agent $i$ is $J$, and its corresponding DPER is denoted as $\mathbb{B}_{J}$.
All agent IDs belonging to type $J$ are denoted as $j \in \mathbb{J}$.
The priority-based sampling strategy in this dataset is denoted as $P_{J}$.
When the data at timestep $t$ of episode $k$ is sampled, the actor $\pi_i$ w.r.t. parameters $\theta_i$ and critic $Q_i$ w.r.t. parameters $\phi_i$ of agent $i$ is trained as follows
\begin{equation} \label{eq_eta}
	\mathcal{L}_{{\rm critic}}(\phi_i)=\mathbb{E}_{P_{J}\left(\mathbb{B}_{J}\right)}
	\left[\frac{\eta}{\hat{u}_{i(j)}^{t,k}} \left(\hat{r}_{i(j)}^{t,k}+ \gamma Q_{i}^{\prime}-Q_{i} \right)^2 \right]
\end{equation}
\begin{equation} \label{eq_beta}
	\mathcal{L}_{{\rm actor}}(\theta_i)=\mathbb{E}_{P_{J}\left(\mathbb{B}_{J}\right)}
	\left[- \frac{\eta}{\hat{u}_{i(j)}^{t,k}} \frac{e^{Q_{i}/\beta}}{Z} \left.Q_{i}\right|_{a=a_{i(j)}^{t,k}} \right]  \,,
\end{equation}
where $(\cdot)_{i(j)}^{(\cdot)}$ indicates that although the original data is sampled from the trajectory of agent $j$, it is used for training the network of agent $i$.
$Q_{i}^{\prime}$ represents target critic.
$e^{Q_{i}/\beta}/Z$ is the filtering trick in CRR, where $Z$ is the normalization coefficient within a mini-batch and $\beta$ is used to control how conservative the policy update is.
The uncertainty $1/\hat{u}_{(\cdot)}^{t,k}$ indicates that policy learning should value individual rewards $\hat{r}_{i(j)}^{t,k}$ that are precisely estimated, since a small $\hat{u}_{(\cdot)}^{t,k}$ means that the reward network has high confidence for the corresponding predicted reward.
$\eta$ is used to control the importance weight of the uncertainty on actor-critic learning.
\begin{figure*}[h]
	\begin{center}
		\includegraphics[width=1\linewidth]{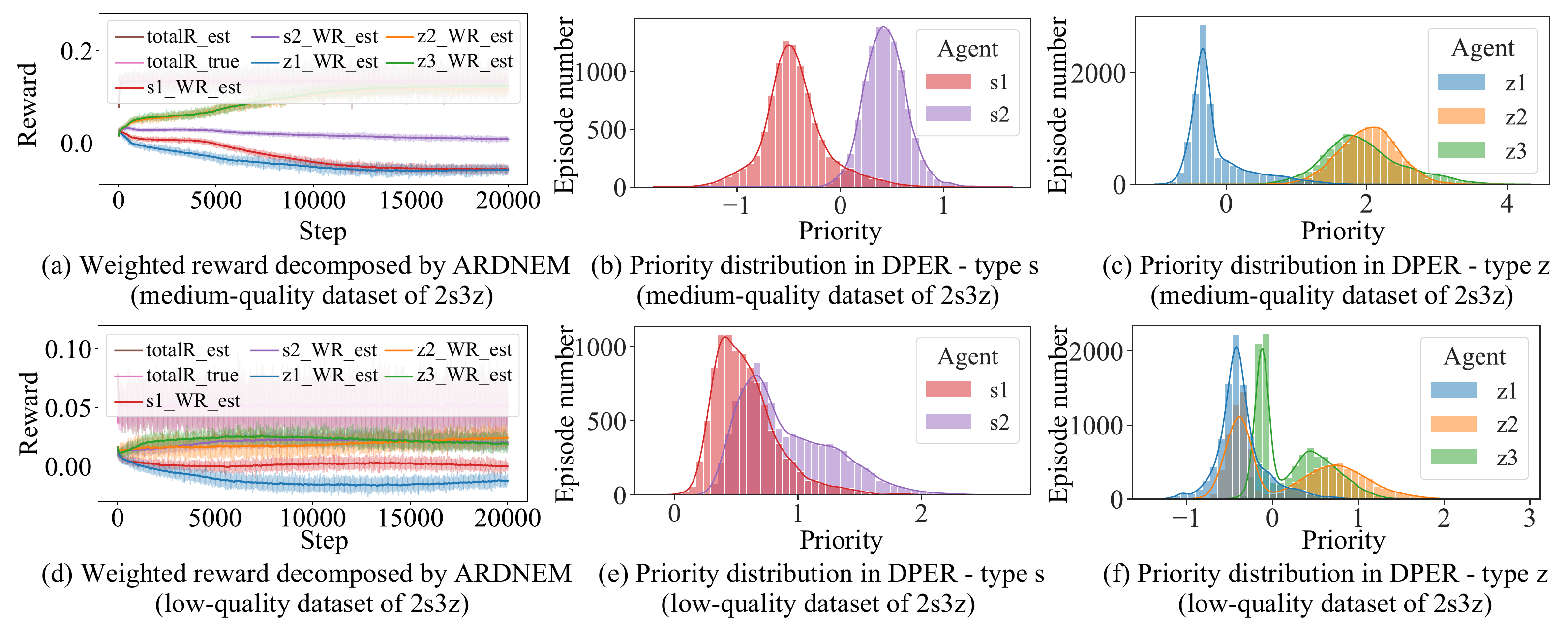}
		\caption{Intermediate results of ARDNEM  and DPER modules on 2s3z datasets.}
		\label{dist}
	\end{center}
\end{figure*}

\section{Experiments}
In this section, we first introduce the data generation method for agent-wise imbalanced multi-agent datasets in StarCraft II \cite{samvelyan2019starcraft}, multi-agent particle environment (MPE) \cite{lowe2017multi} and multi-agent mujoco (MAmujoco) \cite{peng2021facmac}. Then, We evaluate our method SIT on these datasets. Finally, we conduct analytical experiments to better illustrate the superiority of our method.

\subsection{Data Generation for Agent-Wise Imbalanced Multi-Agent Datasets}

We construct the agent-wise imbalanced multi-agent datasets based on six maps in StarCraft II (see Table \ref{SC_dataset_info} in Appendix C.1) and three maps in MPE (see Table \ref{MPE_dataset_info} and Figure \ref{MPE_env} in Appendix C.1) for discrete control, and two maps in MAmujoco (see Figure \ref{MAmujoco_env} in Appendix C.1) for continuous control.
In order to obtain diverse behavior policies for generating these imbalanced datasets, we first online train the joint policies based on QMIX \cite{rashid2018qmix} (discrete) or FacMAC \cite{peng2021facmac} (continuous) in each environment and store them at fixed intervals during the training process. 
Then, these saved joint policies are deposited into random, medium and expert policy pools according to their episode returns, and the details are elaborated in Table \ref{SC_dataset_ER}, Table \ref{MPE_dataset_ER}, Table \ref{MAmujoco_dataset_ER} of Appendix C.2.
Since the policy levels among the individual agents during online training are balanced (\emph{e.g.}, the policy level of each individual agent in a medium joint policy is also medium), we can directly sample the required individual behavior policies from different policy pools to generate the agent-wise imbalanced datasets.

For the convenience of expression, each dataset is represented by the type and policy level corresponding to all individual behavior policies.
\emph{e.g.}, in ${\rm 3s\_vs_\_5z}$ map of StarCraft II, the agent-wise imbalanced multi-agent dataset 100\%$[{\rm s_{1}^{r},s_{2}^{m},s_{3}^{e}}]$ indicates that the policy levels of three Stalkers (agents) that generate this data are random, medium and expert, respectively.
In practice, since we are interested in non-expert data, we generate low-quality and medium-quality agent-wise imbalanced datasets based on the average episode return for all environments. Each of them contains 10000 trajectories and the detailed configuration is shown in Table \ref{SC_dataset_component}, Table \ref{MPE_dataset_component} and Table \ref{MAjoco_dataset_component} of Appendix C.3.

\subsection{Evaluation on Agent-Wise Imbalanced Multi-Agent Datasets}
We compare our proposed method against QMIX (discrete), FacMAC (continuous), behavior cloning (BC), multi-agent version of BCQ \cite{fujimoto2019off} (MABCQ) and CQL \cite{kumar2020conservative} (MACQL), and existing start-of-the-art algorithm ICQ \cite{yang2021believe}. The value decomposition structure of MABCQ and MACQL follows \citet{yang2021believe}. Details for baseline implementations are in Appendix B.
To better demonstrate the effectiveness of our method, we employ the find-tuned hyper-parameters provided by the authors of BCQ and CQL. 
Details for experimental configurations are in Appendix D.  

Table \ref{result} shows the mean and variance of the episodic return of different algorithms with 5 random seeds on tested maps, where the result corresponding to Behavior represents the average episode return of the offline dataset.
It can be found that our method significantly outperforms all baselines in all maps and is even 2x higher than the existing start-of-the-art method ICQ in some maps (\emph{e.g.}, low-quality dataset based on 3s\_vs\_5z in StarCraft II, medium-quality dataset based on CN\_3p1p in MPE and all datasets in MAmujoco), which demonstrates that our method can effectively find and exploit good individual trajectories in agent-wise imbalanced multi-agent datasets to boost the overall performance of multi-agent systems.
Note that since the online algorithm QMIX or FacMAC cannot handle the extrapolation error in offline scenarios, its performance is much lower than other methods.
We also provide the training curves for all datasets in Appendix E.3 to better understand the strengths of our approach.

\begin{figure*}[t]
	\begin{center}
		\includegraphics[width=1\linewidth]{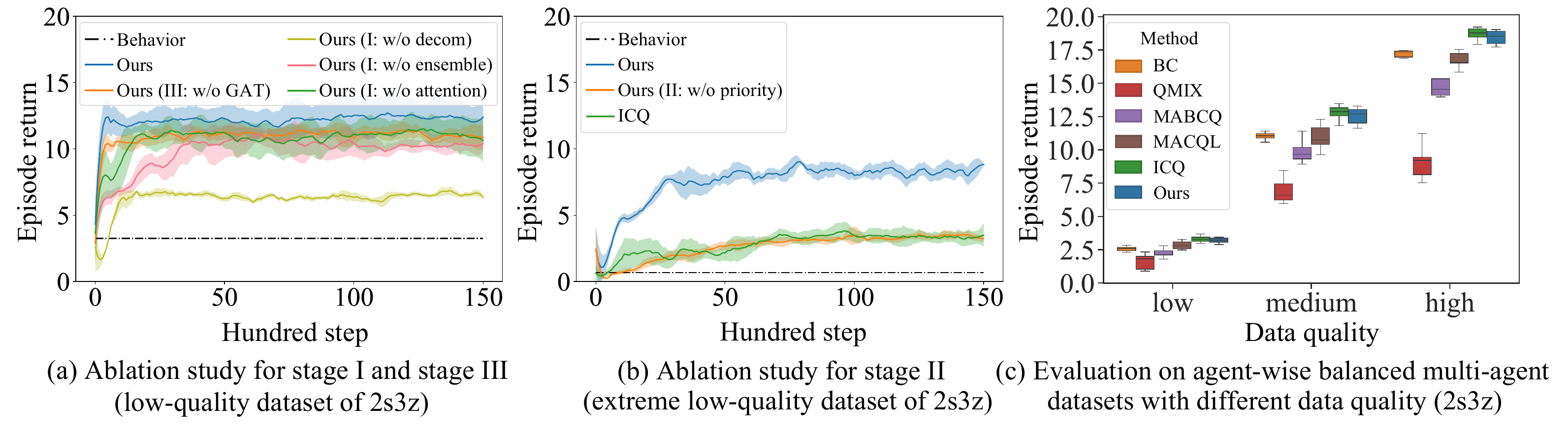}
		\caption{Ablation experiments and the performance on agent-wise balanced datasets.}
		\label{ablation}
	\end{center}
\end{figure*}

\subsection{A Closer Look at SIT}
ARDNEM and DPER are two important modules of our SIT. The former is used to estimate individual rewards, and the latter constructs type-wise prioritized experience replays.
To better demonstrate the effectiveness of our method, we investigate two questions: 1) Can the learned ARDNEM correctly decompose the global reward into individual rewards? and 2) Can the priority in DPER accurately reflect the quality of individual trajectories?
To answer these two questions, we show some intermediate results of our method on medium-quality and low-quality datasets of 2s3z, where the composition of the former is
${\rm 100\% [s_{1}^{r},s_{2}^{e},z_{1}^{r},z_{2}^{e},z_{3}^{e}]}$ and the composition of the latter is
${\rm 50\% [s_{1}^{r},s_{2}^{r},z_{1}^{r},z_{2}^{r},z_{3}^{r}]+ 50\% [s_{1}^{r},s_{2}^{m},z_{1}^{r},z_{2}^{m},z_{3}^{m}]}$. 

Figure \ref{dist}(a) illustrates the decomposed weighted reward $\hat{r}$, estimated by ARDNEM (denoted as WR in the legend) during training on the medium-quality dataset.
We can observe that ${\rm s_1}$'s reward (red) is lower than ${\rm s_2}$'s reward (violet), and ${\rm z_1}$'s reward (blue) is lower than ${\rm z_2}$'s reward (orange) and ${\rm z_3}$'s reward (green).
This decomposition result of ARDNEM agrees with our expectation because all individual trajectories of ${\rm s_1}$ and ${\rm z_1}$ are generated by random behavior policies, while the individual trajectories of other agents are generated by expert behavior policies.
Figure \ref{dist}(d) illustrates the decomposed weighted reward on the low-quality dataset. Similarly, the comparisons also agree with the intended decomposition we desire to achieve.
We further provide the corresponding illustrations for all maps in Appendix E.4, and their decomposition result also matches their data composition (Table \ref{SC_dataset_component}, Table \ref{MPE_dataset_component} and Table \ref{MAjoco_dataset_component} in Appendix C.3).

Since 2s3z map has two types of agents, we store the individual trajectories of each agent into the corresponding prioritized experience replays by their type.
For the medium-quality dataset, Figure \ref{dist}(b) and Figure \ref{dist}(c) show the priority distribution of all individual trajectories in each prioritized experience replay.
It demonstrates that most individual trajectories of ${\rm s_1}$ have lower priority than ${\rm s_2}$, while ${\rm z_1}$ has lower priority than ${\rm z_2}$ and ${\rm z_3}$. This comparison is fully consistent with the quality of the individual trajectories, which indicates the priorities in DPER is as intended.
Figure \ref{dist}(e) and Figure \ref{dist}(f) draw similar conclusions on the low-quality dataset.
A deeper look into Figure \ref{dist}(f) finds that the priority distribution has multiple modals, which indicates that it correctly captures the distribution of the quality even when the variance of the quality is large. \emph{e.g.}, 50\% ${\rm z_2^r}$ + 50\% ${\rm z_2^m}$.

\subsection{Analytical Experiment}

We conduct some analytical experiments to better understand our proposed method. All experiments are evaluated on the low-quality dataset of 2s3z unless otherwise stated.

\paragraph{Ablation study.}
Figure \ref{ablation}(a) illustrates the ablation study about stage I and stage III, where `(w/o decom)' represents that the critic of each agent is directly trained based on the total reward without decomposition.
`(w/o attention)' represents that the weight $\lambda$ of each estimated reward is always equal to 1 instead of a learnable value.
`(w/o ensemble)' removes the ensemble mechanism.
`(w/o GAT)' removes the GAT aggregation.
The result shows that each part is important in our SIT.
To better illustrate the role of priority sampling in stage II, we construct an extreme low-quality dataset based on 2s3z, \emph{i.e.}, ${\rm 99.5\% [s_{1}^{r},s_{2}^{r},z_{1}^{r},z_{2}^{r},z_{3}^{r}]+ 0.5\% [s_{1}^{r},s_{2}^{m},z_{1}^{r},z_{2}^{m},z_{3}^{m}]}$.
The result in Figure \ref{ablation}(b) shows that the priority sampling in stage II plays a key role when the good individual trajectories are sparse in the dataset.

\paragraph{Evaluation on agent-wise balanced multi-agent datasets.}
To evaluate the performance of our method on agent-wise balanced multi-agent datasets, we generate datasets with random, medium and expert quality on the 2s3z for evaluation.
Figure \ref{ablation}(c) shows our method can achieve similar performance to ICQ on these datasets and is significantly higher than other baselines, which indicates that our method can still work well in agent-wise balanced datasets.

\paragraph{Computational complexity.} According to the experiments, the running time of Stage I (supervised learning) is only 10\% of ICQ as it does not contain some complex calculation operations (\emph{e.g.} CRR trick) during gradient backpropagation. The running time of Stage II is negligible (about 10s) as it only needs the inference of the ARDNEM.
The running time of Stage III is similar to ICQ.
Almost all experimental maps can be evaluated within 1 GPU day, except 8m, 1c3s5z and 10m\_vs\_11m maps in StarCraft II which takes 2-3 GPU days.

We also conduct experiments about hyperparameter selection $\alpha,\beta$ and $\eta$. See Appendix E.2 for details. 

\section{Conclusion}
In this paper, we investigate the diversity of individual trajectories in offline multi-agent datasets and empirically show the current offline algorithms cannot fully exploit the useful knowledge in complex and mixed datasets. \emph{e.g.}, agent-wise imbalanced multi-agent datasets.
To address this problem, we propose a novel algorithmic framework called Shared Individual Trajectories (SIT). It can effectively decompose the joint trajectories into individual trajectories, so that the good individual trajectories can be shared among agents to boost the overall performance of multi-agent systems.
Extensive experiments on both discrete control (\emph{i.e.}, StarCraft II and MPE) and continuous control (\emph{i.e.}, MAmujoco) demonstrate the effectiveness and superiority of our method in complex and mixed datasets.

\section{Acknowledgments}
This work was supported in part by National Key Research and Development Program of China (2022YFC3340900), National Natural Science Foundation of China (No. 62037001, U20A20387, No. 62006207), Young Elite Scientists Sponsorship Program by CAST(2021QNRC001), Project by Shanghai AI Laboratory (P22KS00111), the StarryNight Science Fund of Zhejiang University Shanghai Institute for Advanced Study (SN-ZJU-SIAS-0010), Natural Science Foundation of Zhejiang Province (LQ21F020020), Fundamental Research Funds for the Central Universities (226-2022-00142, 226-2022-00051). Baoxiang Wang is partially supported by National Natural Science Foundation of China (62106213, 72150002) and Shenzhen Science and Technology Program (RCBS20210609104356063, JCYJ20210324120011032).

\bibliography{aaai23}

\clearpage
\onecolumn

\section{Appendix} 
\subsection{A. Algorithm} \label{sec_alg}
\begin{algorithm}[h]
	\caption{Shared Individual Trajectories (SIT)}
	\label{alg}
	\KwIn{Offline joint dataset $\mathbb{B}$, training epoch of ARDNEM $L_{{{\rm ARDNEM}}}$, training epoch of actor-critic $L_{{{\rm AC}}}$, priority sampling hyperparameter $\alpha$, conservative hyperparameter $\beta$, uncertainty weight $\eta$, network update interval $d$.}
	\BlankLine
	Initialize ARDNEM $z$ w.r.t. parameters $\psi$, 
	critic networks $Q_i$ w.r.t. parameters $\phi_i$ and actor networks $\pi_i$ w.r.t. parameters $\theta_i$ with random parameters.\\
	Initialize target critic networks $Q_i^{\prime}$.\\
	Initialize DPER $\mathbb{B}_{J}=\emptyset$.\\
	\tcp{Stage I: learning an ARDNEM.}
	\For{$l_{{{\rm ARDNEM}}}=1$ {\rm \bfseries to} $L_{{{\rm ARDNEM}}}$}{
		Sample joint trajectories $\{ o_{1}^{t,k},a_{1}^{t,k},o_{1}^{t+1,k},\dots,o_{{\rm N}}^{t,k},a_{{\rm N}}^{t,k},o_{{\rm N}}^{t+1,k},s^{t,k},r_{tot}^{t,k} \}$ from $\mathbb{B}$. \\
		Train reward decomposition network $z$ with uniform sampling according to 
		$\mathcal{L}_{{\rm MSE}}(\psi)=\mathbb{E}_{{\rm Uniform}(\mathbb{B})} \sum_{m=1}^{M} \frac{1}{M} \left(z(o_{1}^{t,k},a_{1}^{t,k},\dots,o_{{\rm N}}^{t,k},a_{{\rm N}}^{t,k},s^{t,k})-r_{tot}^{t,k}\right)^2$. \\
	}
	\tcp{Stage II: reconstruct the joint offline dataset into type-wise DPER.}
	\For{$k=1$ {\rm \bfseries to} $K$}{
		\For{$i=1$ {\rm \bfseries to} $N$}{
			\For{$t=1$ {\rm \bfseries to} $T$}{
				Obtain ensembled weighted reward $\{\lambda_{i,m}^{t,k} r_{i,m}^{t,k}\}^{m=1,\dots,M}$ according to the input $\{o_{i}^{t,k},a_{i}^{t,k},s^{t,k}\}$ and the learned ARDNEM.\\
				Calculate estimated weighted reward $\hat{r}_{i}^{t,k}=\frac{1}{M} \sum_{m=1}^{M} \lambda_{i,m}^{t,k} r_{i,m}^{t,k}$.\\
				Calculate uncertainty $\hat{u}_i^{t,k}=\sqrt{\frac{1}{M} \sum_{m=1}^{M} \left( \lambda_{i,m}^{t,k} r_{i,m}^{t,k} - \hat{r}_{i}^{t,k} \right)^2}$.\\
				Calculate Monte Carlo return $\hat{g}_i^{t,k}=\sum_{t=t^{\prime}}^{T} \gamma^{t-t^{\prime}} \hat{r}_{i}^{t,k}$.\\
			}
		}
		Calculate episode priority $\hat{p}_i^{k}=\frac{1}{T} \sum_{t=1}^{T} \hat{g}_i^{t,k}$.\\
		Reshape episode priority $p_i^{k}=\frac{e^{\hat{p}_i^{k}/\alpha}}{\sum_{j\in \mathbb{J},k}^{K} e^{\hat{p}_{j}^{k}/\alpha}}$.\\
		Store the individual trajectory $\{\{o_{i}^{t,k},a_{i}^{t,k},o_{i}^{t+1,k},\bm{o_{-i}}^{t,k},\hat{r}_{i}^{t,k},\hat{u}_i^{t,k}  \}^{i=1,\dots,N.t=1,\dots,T},  p_i^{k}\}$ in DPER $\mathbb{B}_{J}$ according to the agent type $J$.
	}
	\tcp{Stage III: conservative policy training with GAT-based critic.}
	\For{$l_{{{\rm AC}}}=1$ {\rm \bfseries to} $L_{{{\rm AC}}}$}{
		Sample the individual trajectory from $\mathbb{B}_{J}$ with priority-based sampling strategy $P_{J}$ for each agent $i$. \\
		Train critic networks $Q_i$ according to 
		$\mathcal{L}_{{\rm critic}}(\phi_i)=\mathbb{E}_{P_{J}\left(\mathbb{B}_{J}\right)}
		\left[\frac{\eta}{\hat{u}_{i(j)}^{t,k}} \left(\hat{r}_{i(j)}^{t,k}+ \gamma Q_{i}^{\prime}\left(\tau_{i(j)}^{t+1,k},a_{i(j)}^{t+1,k},\bm{o_{-i(-j)}}^{t+1,k} \right)-Q_{i}\left( \tau_{i(j)}^{t,k},a_{i(j)}^{t,k},\bm{o_{-i(-j)}}^{t,k}\right) \right)^2 \right]$. \\
		Train actor networks $\pi_i$ according to 
		$\mathcal{L}_{{\rm actor}}(\theta_i)=\mathbb{E}_{P_{J}\left(\mathbb{B}_{J}\right)}
		\left[- \frac{\eta}{\hat{u}_{i(j)}^{t,k}} \frac{e^{Q_{i}\left( \tau_{i(j)}^{t,k},a_{i(j)}^{t,k},\bm{o_{-i(-j)}}^{t,k}\right)/\beta}}{Z} \left.Q_{i}\left( \tau_{i(j)}^{t,k},a,\bm{o_{-i(-j)}}^{t,k}\right)\right|_{a=a_{i(j)}^{t,k}} \right]$. \\
		\If{$l_{{{\rm AC}}}$ \rm mod $d=0$}{
			Update target networks: $Q_i^{\prime}=Q_i$.\\
		}
	}
\end{algorithm}

\subsection{B. Baseline implementations} \label{baseline}
\textbf{BC.} Each agent policy $\pi_i$ w.r.t. parameters $\theta_i$ is optimized by the following loss
\begin{equation}
	\begin{aligned}
		\mathcal{L}_{\rm BC}(\theta_i) = \mathbb{E}_{\tau_i, a_i\sim\mathbb{B}}[-\log(\pi_i(a_i\mid\tau_i))].
	\end{aligned}
\end{equation}

\noindent\textbf{MABCQ.} Suppose the mixer network $Q$ w.r.t. parameters $\phi$, the individual $Q$ networks $Q_i$ w.r.t. parameters $\phi_i$, the behavior policies $\mu_i$ w.r.t. parameters $\theta_i$. We first train behavior policies $\mu_i$ by behavior cloning, then the agent networks is trained by the following loss
\begin{equation}
	\begin{aligned}
		\mathcal{L}_{\rm BCQ}(\phi,\phi_i) & =\mathbb{E}_{\bm{\tau},\bm{a},r,\bm{\tau^{\prime}}\sim \mathbb{B}}\left[
		\left(r + \gamma \max_{\tilde{\bm{a}}^{[j]}}Q^{\prime}(\bm{\tau^{\prime}}, \tilde{\bm{a}}^{[j]}) - Q(\bm{\tau}, \bm{a})
		\right)^2 \right]\\
		& \qquad \quad \quad \tilde{\bm{a}}^{[j]} = \bm{a}^{[j]} + \xi(\bm{\tau}, \bm{a}^{[j]})
	\end{aligned},
\end{equation}
where $Q(\bm{\tau}, \bm{a}) = w_i Q_i(\tau_i,a_i) + b$ and $\xi(\bm{\tau}, \bm{a}^{[j]})$ denotes the perturbation model, which is decomposed as $\xi_i(\tau_i, a_i^{[j]})$.
If $\frac{a_i^{[j]} \sim \mu_i(\tau_i)}{\max\{a_i^{[j]} \sim \mu_i(\tau_i)\}_{j=1}^{m} } \leq \zeta$ in agent $i$, $a_i^{[j]}$ is considered an unfamiliar action and $\xi^i(\tau^i, a_i^{[j]})$ will mask $a_i^{[j]}$ in maximizing $Q_i$-value operation.

\noindent\textbf{MACQL.} Suppose the mixer network $Q$ w.r.t. parameters $\phi$, the individual $Q$ networks $Q_i$ w.r.t. parameters $\phi_i$, then the agent networks is trained by the following loss
\begin{equation}
	\begin{aligned}
		\mathcal{L}_{\rm CQL}(\phi,\phi_i) & = \beta_{\rm CQL}\mathbb{E}_{\tau_i,a_i,\bm{\tau},\bm{a}\sim\mathbb{B}}\left[
		\sum_i \log\sum_{a_i}\exp(w_iQ_i(\tau_i,a_i) + b) - \mathbb{E}_{ \bm{a}\sim\bm{\mu}(\bm{a}\mid\bm{\tau})}[Q(\bm{\tau}, \bm{a})]\right] \\
		& \qquad \qquad \qquad \qquad \qquad \qquad \qquad \qquad + \frac{1}{2}\mathbb{E}_{\bm{\tau}\sim\mathbb{B}, \bm{a}\sim\bm{\mu}(\bm{a}\mid\bm{\tau})}\left[
		\left(y_{\rm CQL} - Q(\bm{\tau}, \bm{a})\right)^2\right], \\
	\end{aligned}
\end{equation}
where $y_{\rm CQL}$ is calculated based on $n$-step off-policy estimation~(e.g., Tree Backup algorithm).

\noindent\textbf{ICQ.} Suppose the centralized critic network $Q$ w.r.t. parameters $\phi$, the individual critic networks $Q_i$ w.r.t. parameters $\phi_i$, the individual policies $\pi_i$ w.r.t. parameters $\theta_i$, then the actor-critic is trained by the following loss
\begin{equation}
	\begin{aligned}
		\mathcal{L}_{{\rm actor}}(\theta_i) & = \mathbb{E}_{\tau_i, a_i\sim \mathbb{B}}\left[-\frac{1}{Z_i}\log(\pi_i(a_i\mid\tau_i))\exp\left(\frac{w_i Q_{i}(\tau_i, a_i)}{\beta_{{\rm ICQ}}} \right)\right] \\
		\mathcal{L}_{{\rm critic}}(\phi, \phi_i) & =\mathbb{E}_{\bm{\tau},\bm{a},r,\bm{\tau^{\prime}},\bm{a^{\prime}} \sim \mathbb{B}}\left[\sum_{t\geq 0}(\gamma\lambda_{{\rm ICQ}})^t\left[
		r + \gamma
		\frac{\exp\left(\frac{1}{\beta_{{\rm ICQ}}}Q(\bm{\tau^{\prime}},\bm{a^{\prime}})\right)}{Z}
		Q(\bm{\tau^{\prime}},\bm{a^{\prime}})-Q(\bm{\tau}, \bm{a})\right]
		\right]^2, \\
	\end{aligned}
\end{equation}
where $Q(\bm{\tau}, \bm{a}) = w_i Q_i(\tau_i,a_i) + b$ represents the linear mixer of the centralized critic network.

\subsection{C. Environment and offline dataset information} \label{app_dataset}
\noindent\textbf{C.1 Environment information} \label{dataset_info} 

\noindent In this paper, we construct the agent-wise imbalanced multi-agent datasets based on six maps in StarCraft II, three modified maps for partially observable settings in the multi-agent particle environment (MPE) and two maps in multi-agent mujoco (MAmujoco). To better understand our experiments, we give a brief introduction to these environments.
\begin{table}[h]
	\small
	\centering
	\begin{tabular}{cccc}
		\toprule
		Maps         & Controlled ally agents              & Built-in enemy agents               & Difficulties \\ \midrule
		2s\_vs\_1sc  & 2 Stalkers                          & 1 Spine Crawler                     & Super Hard   \\ \specialrule{0em}{1pt}{1pt}
		3s\_vs\_5z   & 3 Stalkers                          & 5 Zealots                           & Super Hard   \\ \specialrule{0em}{1pt}{1pt}
		2s3z         & 2 Stalkers \& 3 Zealots             & 2 Stalkers \& 3 Zealots             & Super Hard   \\ \specialrule{0em}{1pt}{1pt}
		8m           & 8 Marines                           & 8 Marines                           & Super Hard   \\ \specialrule{0em}{1pt}{1pt}
		1c3s5z       & 1 Colossus, 3 Stalkers \& 5 Zealots & 1 Colossus, 3 Stalkers \& 5 Zealots & Super Hard   \\ \specialrule{0em}{1pt}{1pt}
		10m\_vs\_11m & 10 Marines                          & 11 Marines                          & Super Hard   \\ \bottomrule
	\end{tabular}
	\vspace{6pt}
	\caption{The information of test maps on StarCraft II.} 
	\label{SC_dataset_info}
\end{table}

\paragraph{StarCraft II.} StarCraftII micromanagement aims to accurately control the individual units and complete cooperation tasks. 
In this environment, each enemy unit is controlled by the built-in AI while each allied unit is controlled by a learned policy. 
The local observation of each agent contains the following attributes for both allied and enemy units within the sight range: ${\rm distance}$, ${\rm relative\_x}$, ${\rm relative\_ y}$, ${\rm health}$, ${\rm shield}$, and ${\rm unit\_type}$. 
The action space of each agent includes:
${\rm noop}$, ${\rm move[direction]}$, ${\rm attack[enemy id]}$, and ${\rm stop}$.
Under the control of these actions, each agent can move and attack in various maps.
The reward setting is based on the hit-point damage dealt on the enemy units, together with special bonuses for killing the enemy units and winning the battle. 
In our experiments, we choose 2s\_vs\_1sc, 3s\_vs\_5z, 2s3z, 8m, 1c3s5z and 10m\_vs\_11m for evaluation, and more detailed information about these maps is shown in Table \ref{SC_dataset_info}.

\begin{figure}[h]
	\begin{center}
		\includegraphics[width=0.99\linewidth]{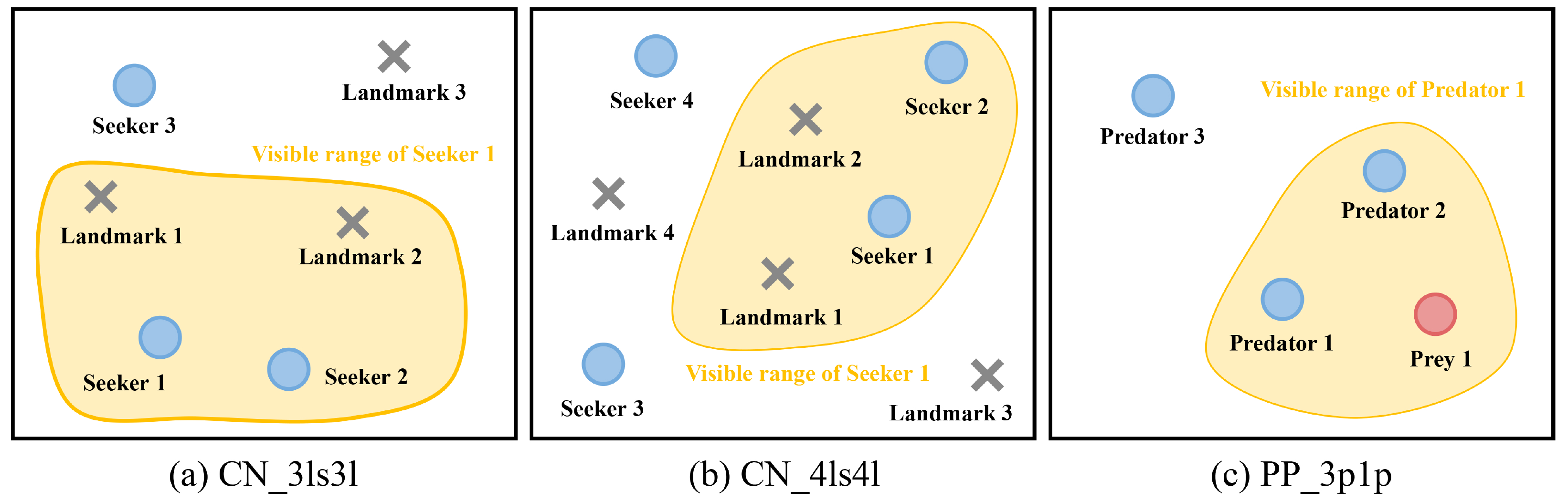}
		\caption{Three modified maps for partially observable settings in MPE.}
		\label{MPE_env}
	\end{center}
\end{figure}

\begin{table}[h]
	\small
	\centering
	\begin{tabular}{ccccc}
		\toprule
		Maps                           & Controlled agents  & Targets & Visible agents & Visible targets\\ \midrule
		\begin{tabular}[c]{@{}c@{}}${\text{CN\_3ls3l}}$\\ $\text{(Cooperative navigation)}$ \end{tabular}  & 3 Landmark Seekers & 3 Landmarks               & 1 Landmark Seeker & 2 Landmarks     \\ \specialrule{0em}{1pt}{1pt}
		\begin{tabular}[c]{@{}c@{}}${\text{CN\_4ls4l}}$\\ $\text{(Cooperative navigation)}$ \end{tabular}  & 4 Landmark Seekers & 4 Landmarks               & 1 Landmark Seeker & 2 Landmarks      \\ \specialrule{0em}{1pt}{1pt}
		\begin{tabular}[c]{@{}c@{}}${\text{PP\_3p1p}}$\\ $\text{(Predator-prey)}$ \end{tabular}            & 3 Predators        & 1 Prey          & 1 Predator  & 1 Prey             \\ \bottomrule
	\end{tabular}
	\vspace{6pt}
	\caption{The information of test maps on MPE.} 
	\label{MPE_dataset_info}
\end{table}

\paragraph{Multi-agent particle environment (MPE).} MPE is a common benchmark that contains various multi-agent games.  In this paper, we choose the discrete version of cooperative navigation (CN) and predator-prey (PP) for evaluation.
In CN task, landmark seeks (agents) must cooperate through physical actions to reach a set of landmarks without colliding with each other.
In PP task, some slower cooperating predators must chase the faster prey and avoid collisions in a randomly generated environment. We set the prey as a heuristic agent whose movement direction is always away from the nearest predator.
Note that in the original MPE environment, each agent can directly observe the global information of the environment. However, we modify the environment to the partially observable setting, a more challenging and practical setting, thus we reset the local observation of the agents in each map as shown in Figure \ref{MPE_env} and Table \ref{MPE_dataset_info}.

\begin{figure}[h]
	\begin{center}
		\includegraphics[width=0.5\linewidth]{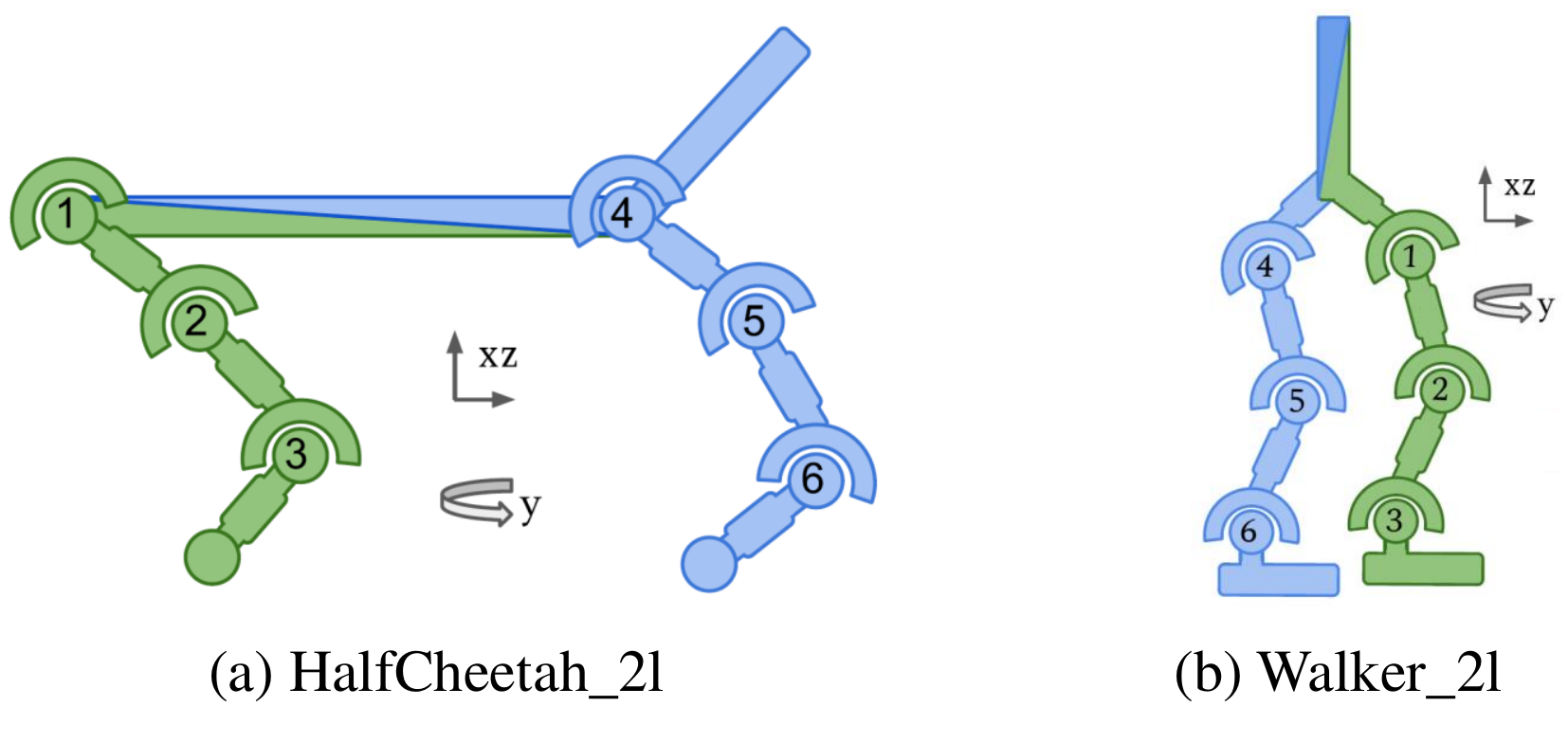}
		\caption{Two maps in MAmujoco.}
		\label{MAmujoco_env}
	\end{center}
\end{figure}

\paragraph{Multi-agent mujoco (MAmujoco).} MAmujoco is a novel benchmark for continuous cooperative multi-agent robotic control. Starting from the popular single-agent robotic mujoco \cite{todorov2012mujoco} control suite included with OpenAI Gym \cite{brockman2016openai}. Each agent’s action space in MAmujoco is given by the joint action space over all motors controllable by that agent. 
As shown in Figure \ref{MAmujoco_env}, we select HalfCheetah\_2l and Walker\_2l maps with 200 termination steps for evaluation. 
In these two maps, each agent can control joints of one color. For example, the agent corresponding to the green partition in HalfCheetah\_2l (Figure \ref{MAmujoco_env}(a)) consists of three joints (joint ids 1, 2, and 3) and four adjacent body segments. Each joint has an action space $[-1,1]$, so the action space for each agent is a 3-dimensional vector with each entry in $[-1,1]$.

\ \newline
\noindent\textbf{C.2 Policy pools for each environments} \label{dataset_ER}

\noindent In order to obtain diverse behavior policies for generating agent-wise imbalanced multi-agent datasets, we first online train the joint policies based on QMIX \cite{rashid2018qmix} (discrete) or FacMAC \cite{peng2021facmac} (continuous) in each environment and store them at fixed intervals during the training process. 
Then, these saved joint policies are deposited into random, medium and expert policy pools according to their episode returns.
Table \ref{SC_dataset_ER}, Table \ref{MPE_dataset_ER} and Table \ref{MAmujoco_dataset_ER} show the episode return corresponding to different policy pools in each map, where `Episode return range' is determined by fully random policies and the best policies trained by QMIX or FacMAC.
After these policy pools are constructed, we can directly sample the required individual behavior policies from different policy pools to generate the agent-wise imbalanced multi-agent datasets.
\begin{table}[h]
	\small
	\centering
	\begin{tabular}{ccccc}
		\toprule
		Maps         & \begin{tabular}[c]{@{}c@{}}Random\\ pools \end{tabular} & \begin{tabular}[c]{@{}c@{}}Medium\\ pools \end{tabular} & \begin{tabular}[c]{@{}c@{}}Expert\\ pools \end{tabular} & \begin{tabular}[c]{@{}c@{}}Episode\\ return range \end{tabular}\\ \midrule
		2s\_vs\_1sc  & 0$\sim$6           & 6$\sim$14          & 14$\sim$20         & 0$\sim$20               \\ \specialrule{0em}{1pt}{1pt}
		3s\_vs\_5z   & 0$\sim$6           & 6$\sim$14          & 14$\sim$20         & 0$\sim$20             \\ \specialrule{0em}{1pt}{1pt}
		2s3z         & 0$\sim$6           & 6$\sim$14          & 14$\sim$20         & 0$\sim$20            \\ \specialrule{0em}{1pt}{1pt}
		8m           & 0$\sim$6           & 6$\sim$14          & 14$\sim$20         & 0$\sim$20             \\ \specialrule{0em}{1pt}{1pt}
		1c3s5z       & 0$\sim$6           & 6$\sim$14          & 14$\sim$20         & 0$\sim$20            \\ \specialrule{0em}{1pt}{1pt}
		10m\_vs\_11m & 0$\sim$6           & 6$\sim$14          & 14$\sim$20         & 0$\sim$20             \\ \bottomrule
	\end{tabular}
	\vspace{6pt}
	\caption{Episode return of different policy pools in StarCraft II.} 
	\label{SC_dataset_ER}
\end{table}

\begin{table}[h]
	\small
	\centering
	\begin{tabular}{ccccc}
		\toprule
		Maps         & \begin{tabular}[c]{@{}c@{}}Random\\ pools \end{tabular} & \begin{tabular}[c]{@{}c@{}}Medium\\ pools \end{tabular} & \begin{tabular}[c]{@{}c@{}}Expert\\ pools \end{tabular} & \begin{tabular}[c]{@{}c@{}}Episode\\ return range \end{tabular}\\ \midrule
		CN\_3ls3l & -174.41$\sim$-133.28 & -133.28$\sim$-78.44  & -78.44$\sim$-37.32  & -174.41$\sim$-37.32   \\ \specialrule{0em}{1pt}{1pt}
		CN\_4ls4l & -288.89$\sim$-214.47 & -214.47$\sim$-115.24 & -115.24$\sim$-40.81 & -288.89$\sim$-40.81   \\ \specialrule{0em}{1pt}{1pt}
		PP\_3p1p           & -306.57$\sim$-231.08 & -231.08$\sim$-130.44 & -130.44$\sim$-54.96 & -306.57$\sim$-54.96   \\ \bottomrule
	\end{tabular}
	\vspace{6pt}
	\caption{Episode return of different policy pools in MPE.} 
	\label{MPE_dataset_ER}
\end{table}

\begin{table}[h]
	\small
	\centering
	\begin{tabular}{ccccc}
		\toprule
		Maps         & \begin{tabular}[c]{@{}c@{}}Random\\ pools \end{tabular} & \begin{tabular}[c]{@{}c@{}}Medium\\ pools \end{tabular} & \begin{tabular}[c]{@{}c@{}}Expert\\ pools \end{tabular} & \begin{tabular}[c]{@{}c@{}}Episode\\ return range \end{tabular}\\ \midrule
		HalfCheetah\_2l & -203.47$\sim$-69.42 & -69.42$\sim$64.63  & 64.63$\sim$198.70  & -203.47$\sim$198.70   \\ \specialrule{0em}{1pt}{1pt}
		Walker\_2l           & -74.51$\sim$27.35 & 27.35$\sim$129.21 & 129.21$\sim$231.09 & -74.51$\sim$231.09   \\ \bottomrule
	\end{tabular}
	\vspace{6pt}
	\caption{Episode return of different policy pools in MAmujoco.} 
	\label{MAmujoco_dataset_ER}
\end{table}

\ \newline
\noindent\textbf{C.3 Data components of agent-wise imbalanced offline multi-agent datasets} \label{dataset_component}

\noindent After the policy pools are ready, we sample the individual behavior policies from these pools according to the data composition in Table \ref{SC_dataset_component}, Table \ref{MPE_dataset_component} and Table \ref{MAjoco_dataset_component}, and generate agent-wise imbalanced offline datasets. Note that since we are interested in non-expert data, We mainly generate two types of datasets. \emph{i.e.}, low-quality and medium-quality datasets.
\begin{table}[h]
	\small
	\centering
	\begin{tabular}{ccccc}
		\toprule
		& Maps         & Components & \begin{tabular}[c]{@{}c@{}}Average\\ episode return \end{tabular}   & \begin{tabular}[c]{@{}c@{}}Episode\\ number \end{tabular}\\ \midrule
		\multirow{6}{*}{\vspace{-55pt}\shortstack{Low\\Quality}} & 2s\_vs\_1sc  & \begin{tabular}[c]{@{}l@{}}${\rm 50\% [s_{1}^{r},s_{2}^{r}]+}$\\ ${\rm 50\% [s_{1}^{r},s_{2}^{m}]}$ \end{tabular}          & 2.84 & 10000               \\ \specialrule{0em}{1pt}{1pt}
		& 3s\_vs\_5z   & \begin{tabular}[c]{@{}l@{}}${\rm 50\% [s_{1}^{r},s_{2}^{r},s_{3}^{r}]+}$\\ ${\rm 50\% [s_{1}^{r},s_{2}^{r},s_{3}^{m}]}$ \end{tabular}          & 2.94& 10000               \\ \specialrule{0em}{1pt}{1pt}
		& 2s3z         & \begin{tabular}[c]{@{}l@{}}${\rm 50\% [s_{1}^{r},s_{2}^{r},z_{1}^{r},z_{2}^{r},z_{3}^{r}]+}$\\ ${\rm 50\% [s_{1}^{r},s_{2}^{m},z_{1}^{r},z_{2}^{m},z_{3}^{m}]}$ \end{tabular}          & 3.24& 10000               \\ \specialrule{0em}{1pt}{1pt}
		& 8m           & \begin{tabular}[c]{@{}l@{}}${\rm 50\% [m_{1}^{r},m_{2}^{r},m_{3}^{r},m_{4}^{r},m_{5}^{r},m_{6}^{r},m_{7}^{r},m_{8}^{r}]+}$\\ ${\rm 50\% [m_{1}^{r},m_{2}^{r},m_{3}^{r},m_{4}^{r},m_{5}^{r},m_{6}^{r},m_{7}^{r},m_{8}^{m}]}$ \end{tabular}          & 3.10& 10000               \\ \specialrule{0em}{1pt}{1pt}
		& 1c3s5z       & \begin{tabular}[c]{@{}l@{}}${\rm 50\% [c_{1}^{r},s_{1}^{r},s_{2}^{r},s_{3}^{r},z_{1}^{r},z_{2}^{r},z_{3}^{r},z_{4}^{r},z_{5}^{r}]+}$\\ ${\rm 50\% [c_{1}^{m},s_{1}^{r},s_{2}^{r},s_{3}^{m},z_{1}^{r},z_{2}^{r},z_{3}^{r},z_{4}^{r},z_{5}^{m}]}$ \end{tabular}          & 5.56& 10000               \\ \specialrule{0em}{1pt}{1pt}
		& 10m\_vs\_11m & \begin{tabular}[c]{@{}l@{}}${\rm 50\% [m_{1}^{r},m_{2}^{r},m_{3}^{r},m_{4}^{r},m_{5}^{r},m_{6}^{r},m_{7}^{r},m_{8}^{r},m_{9}^{r},m_{10}^{r}]+}$\\ ${\rm 50\% [m_{1}^{r},m_{2}^{r},m_{3}^{m},m_{4}^{m},m_{5}^{m},m_{6}^{m},m_{7}^{m},m_{8}^{m},m_{9}^{m},m_{10}^{m}]}$ \end{tabular}          & 3.82& 10000               \\ \midrule
		\multirow{6}{*}{\vspace{-10pt}\shortstack{Medium\\Quality}} & 2s\_vs\_1sc  & ${\rm 100\% [s_{1}^{r},s_{2}^{e}]}$          & 9.86& 10000               \\ \specialrule{0em}{1pt}{1pt}
		& 3s\_vs\_5z   & ${\rm 100\% [s_{1}^{r},s_{2}^{r},s_{3}^{e}]}$          & 7.06& 10000               \\ \specialrule{0em}{1pt}{1pt}
		& 2s3z         & ${\rm 100\% [s_{1}^{r},s_{2}^{e},z_{1}^{r},z_{2}^{e},z_{3}^{e}]}$          & 7.04& 10000               \\ \specialrule{0em}{1pt}{1pt}
		& 8m           & ${\rm 100\% [m_{1}^{r},m_{2}^{e},m_{3}^{e},m_{4}^{e},m_{5}^{e},m_{6}^{e},m_{7}^{e},m_{8}^{e}]}$          & 9.85& 10000               \\ \specialrule{0em}{1pt}{1pt}
		& 1c3s5z       & ${\rm 100\% [c_{1}^{e},s_{1}^{r},s_{2}^{r},s_{3}^{e},z_{1}^{r},z_{2}^{r},z_{3}^{r},z_{4}^{r},z_{5}^{e}]}$          & 10.31& 10000              \\ \specialrule{0em}{1pt}{1pt}
		& 10m\_vs\_11m & ${\rm 100\% [m_{1}^{r},m_{2}^{e},m_{3}^{e},m_{4}^{e},m_{5}^{e},m_{6}^{e},m_{7}^{e},m_{8}^{e},m_{9}^{e},m_{10}^{e}]}$          & 9.23& 10000               \\ \bottomrule
	\end{tabular}
	\vspace{6pt}
	\caption{Data components of agent-wise imbalanced offline multi-agent datasets in StarCraft II.} 
	\label{SC_dataset_component}
\end{table}

\begin{table}[h]
	\small
	\centering
	\begin{tabular}{ccccc}
		\toprule
		& Maps         & Components & \begin{tabular}[c]{@{}c@{}}Average\\ episode return \end{tabular}   & \begin{tabular}[c]{@{}c@{}}Episode\\ number \end{tabular}\\ \midrule
		\multirow{6}{*}{\vspace{6pt}\shortstack{Random\\Quality}} & CN\_3ls3l & \begin{tabular}[c]{@{}l@{}}${\rm 50\% [ls_{1}^{r},ls_{2}^{r},ls_{3}^{r}]+}$\\ ${\rm 50\% [ls_{1}^{r},ls_{2}^{r},ls_{3}^{m}]}$ \end{tabular}          & -157.70&10000 \\ \specialrule{0em}{1pt}{1pt}
		& CN\_4ls4l   & \begin{tabular}[c]{@{}l@{}}${\rm 50\% [ls_{1}^{r},ls_{2}^{r},ls_{3}^{r},ls_{4}^{r}]+}$\\ ${\rm 50\% [ls_{1}^{r},ls_{2}^{r},ls_{3}^{r},ls_{4}^{m}]}$ \end{tabular}          & -278.40&10000               \\ \specialrule{0em}{1pt}{1pt}
		& PP\_3p1p         & \begin{tabular}[c]{@{}l@{}}${\rm 50\% [p_{1}^{r},p_{2}^{r},p_{3}^{r}]+}$\\ ${\rm 50\% [p_{1}^{r},p_{2}^{r},p_{3}^{m}]}$ \end{tabular}          & -249.58&10000               \\ \midrule
		\multirow{6}{*}{\vspace{23pt}\shortstack{Medium\\Quality}} & CN\_3ls3l  & ${\rm 100\% [ls_{1}^{r},ls_{2}^{m},ls_{3}^{e}]}$          & -107.46&10000               \\ \specialrule{0em}{1pt}{1pt}
		& CN\_4ls4l   & ${\rm 100\% [ls_{1}^{r},ls_{2}^{r},ls_{3}^{e},ls_{4}^{e}]}$          & -166.26&10000               \\ \specialrule{0em}{1pt}{1pt}
		& PP\_3p1p         & ${\rm 100\% [p_{1}^{r},p_{2}^{m},p_{3}^{e}]}$          & -155.46&10000               \\ \bottomrule
	\end{tabular}
	\vspace{6pt}
	\caption{Data components of agent-wise imbalanced offline multi-agent datasets in MPE.} 
	\label{MPE_dataset_component}
\end{table}

\begin{table}[h]
	\small
	\centering
	\begin{tabular}{ccccc}
		\toprule
		& Maps         & Components & \begin{tabular}[c]{@{}c@{}}Average\\ episode return \end{tabular}   & \begin{tabular}[c]{@{}c@{}}Episode\\ number \end{tabular}\\ \midrule
		\multirow{6}{*}{\vspace{36pt}\shortstack{Random\\Quality}} & HalfCheetah\_2l & \begin{tabular}[c]{@{}l@{}}${\rm 100\% [l_{1}^{m},l_{2}^{r}]}$ \end{tabular}          & -110.52&10000 \\ \specialrule{0em}{1pt}{1pt}
		& Walker\_2l         & \begin{tabular}[c]{@{}l@{}}${\rm 100\% [l_{1}^{r},l_{2}^{m}]}$ \end{tabular}          & -21.67&10000               \\ \midrule
		\multirow{6}{*}{\vspace{35pt}\shortstack{Medium\\Quality}} & HalfCheetah\_2l  & ${\rm 100\% [l_{1}^{r},l_{2}^{e}]}$          & 41.79&10000               \\ \specialrule{0em}{1pt}{1pt}
		& Walker\_2l         & ${\rm 100\% [l_{1}^{r},l_{2}^{e}]}$          & 71.68&10000               \\ \bottomrule
	\end{tabular}
	\vspace{6pt}
	\caption{Data components of agent-wise imbalanced offline multi-agent datasets in MAmujoco.} 
	\label{MAjoco_dataset_component}
\end{table}

\subsection{D. Experimental configurations} \label{exper_config}

\noindent\textbf{D.1 Hardware configurations}

\noindent In all experiments, the GPU is NVIDIA A100 and the CPU is AMD EPYC 7H12 64-Core Processor.

\ \newline
\noindent\textbf{D.2 Hyperparameters}

\noindent In our proposed algorithmic framework SIT, we first train ARDNEM with the hyperparameters shown in Figure \ref{hyper_ARDNEM}, then take the last checkpoint to construct type-wise DPER, and finally use the hyperparameters shown in Figure \ref{hyper_agent} to train the agent policies. To ensure a fair comparison, other baselines also share most hyperparameters with ours during agent training as shown in Figure \ref{hyper_agent}.

\clearpage

\begin{minipage}[c]{0.5\textwidth}
	\centering
	\begin{tabular}{ll}
		\toprule
		Hyperparameter & Value \\
		\midrule
		\hspace{0.3cm} Learning rate & $1 \times 10^{-4}$\\
		\hspace{0.3cm} Optimizer & RMS \\
		\hspace{0.3cm} Training epoch & 20000 \\
		\hspace{0.3cm} Gradient clipping & 10 \\
		\hspace{0.3cm} Reward network dimension & 64 \\
		\hspace{0.3cm} Attention hidden dimension & 64 \\
		\hspace{0.3cm} Activation function & ReLU \\
		\hspace{0.3cm} Batch size & 32 \\
		\hspace{0.3cm} Replay buffer size & $1.0\times 10^4$ \\
		\hspace{0.3cm} Ensemble number & 5 \\
		\bottomrule
	\end{tabular}
	\vspace{-6pt}
	\captionof{table}{Hyperparameters sheet for ARDNEM learning\label{hyper_ARDNEM}}
\end{minipage}
\begin{minipage}[h]{0.5\textwidth}
	\centering
	\begin{tabular}{ll}
		\toprule
		Hyperparameter & Value \\
		\midrule
		Shared& \\
		\hspace{0.3cm} Agent network learning rate & $5 \times 10^{-4}$\\
		\hspace{0.3cm} Optimizer & RMS \\
		\hspace{0.3cm} Discount factor & 0.99 \\
		\hspace{0.3cm} Training epoch & 15000 \\
		\hspace{0.3cm} Parameters update interval & 100 \\
		\hspace{0.3cm} Gradient clipping & 10 \\
		\hspace{0.3cm} Mixer/Critic hidden dimension & 32 \\
		\hspace{0.3cm} RNN hidden dimension & 64 \\
		\hspace{0.3cm} Activation function & ReLU \\
		\hspace{0.3cm} Batch size & 32 \\
		\hspace{0.3cm} Replay buffer size & $1.0\times 10^4$ \\
		\midrule
		MABCQ & \\
		\hspace{0.3cm} $\zeta$ & 0.3 \\
		MACQL & \\
		\hspace{0.3cm} $\beta_{\rm CQL}$ & 2.0 \\
		ICQ & \\
		\hspace{0.3cm} Critic network learning rate & $1 \times 10^{-4}$\\
		\hspace{0.3cm} $\beta_{\rm ICQ}$ & 0.1 \\
		\hspace{0.3cm} $\lambda_{\rm ICQ}$ & 0.8 \\
		Ours & \\
		\hspace{0.3cm} Critic network learning rate & $1 \times 10^{-4}$\\
		\hspace{0.3cm} $\alpha$ & 0.2 \\
		\hspace{0.3cm} $\beta$ & 0.1 \\
		\hspace{0.3cm} $\eta$ & 1 \\
		\hspace{0.3cm} Rescaled priority range & [0,20] \\
		\bottomrule
	\end{tabular}
	\vspace{-6pt}
	\captionof{table}{Hyperparameters sheet for agent learning\label{hyper_agent}}
\end{minipage}

\subsection{E. Experiments}

\noindent\textbf{E.1 Reward decomposition network with the monotonic nonlinear constraint (MNC).}

\begin{figure}[h]
	\begin{center}
		\includegraphics[width=\linewidth]{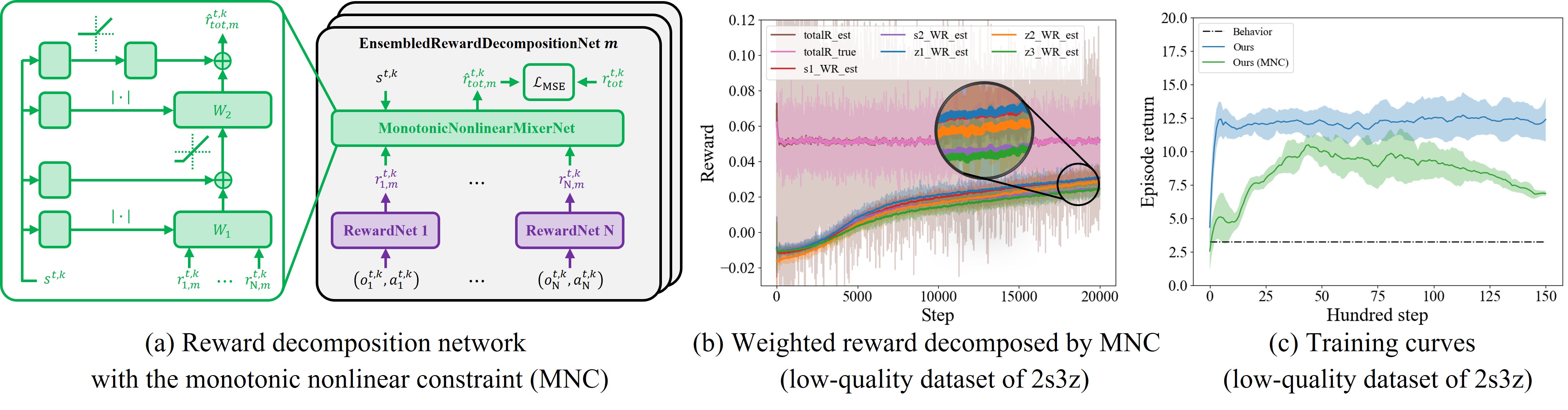}
		\caption{Experiment about reward decomposition network with the monotonic nonlinear constraint (MNC).}
		\label{non}
	\end{center}
\end{figure}

\noindent In our SIT, we propose an attention-based reward decomposition network in the stage I. A natural alternative is the decomposition module with monotonic nonlinear constraint (MNC),  similar to QMIX \cite{rashid2018qmix}. 
Specifically, we only need to use the hypernetwork shown in Figure \ref{non}(a), so that the total reward $r_{tot}$ and the individual reward $r_{i}$ satisfy the following relationship to achieve reward decomposition
\begin{equation}
	\frac{\partial r_{tot}}{\partial r_i}  \geq 0,~ \forall i \in N.
	\label{monotonicity}
\end{equation}

Figure \ref{non}(b) shows the reward decomposition result of MNC. We can observe 
there are two problems with the decomposition result. 1) The gap between individual rewards is small, so the quality of the trajectory cannot be well distinguished. 2) More importantly, the reward decomposition result is incorrect. According to the data composition of low-quality dataset of 2s3z. \emph{i.e.}, Table \ref{SC_dataset_component}. we expect s1 (red)$<$s2 (violet), z1 (blue)$<$z2 (orange), z3 (green), Figure \ref{non}(b) does not match this expectation. Instead, the decomposition result of our SIT is as intended. \emph{i.e.}, Figure \ref{dist}(d).
Therefore, policy learning with MNC-decomposed rewards cannot achieve the satisfactory result, as shown in Figure \ref{non}(c).

\ \newline
\noindent\textbf{E.2 Hyperparameter selection $\alpha,\beta$ and $\eta$.}

\begin{figure}[h]
	\begin{center}
		\includegraphics[width=0.885\linewidth]{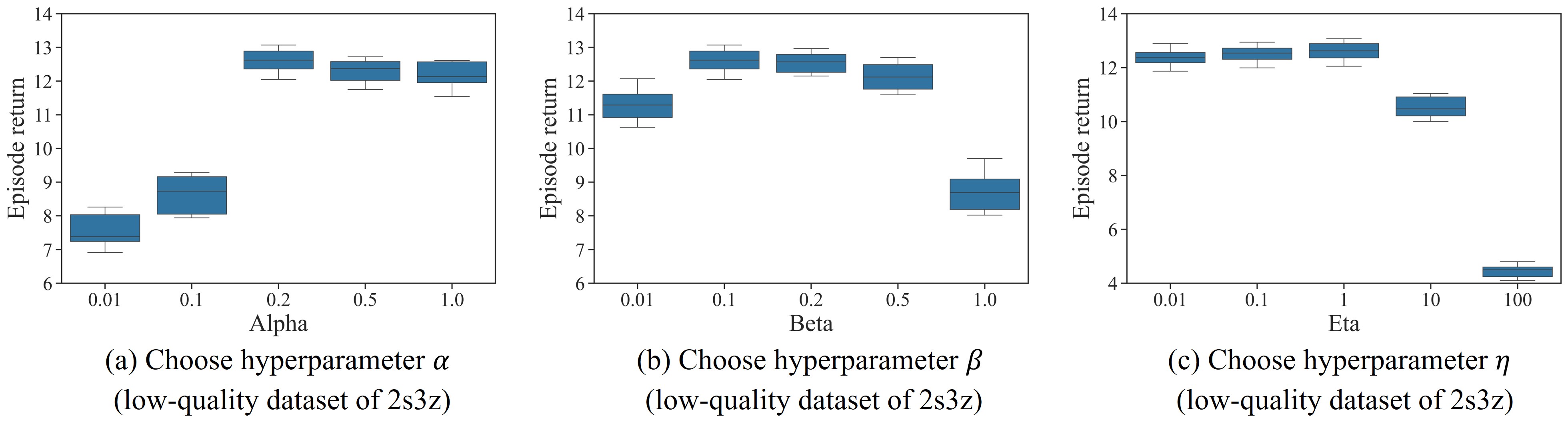}
		\caption{Hyperparameter selection.}
		\label{hyper}
	\end{center}
\end{figure}

\noindent The hyperparameter $\alpha$ in Equation \eqref{eq_alpha} determines how much prioritization is used during sampling training data.
If $\alpha \rightarrow 0$, it means we only sample the highest quality individual trajectory,
while $\alpha \rightarrow \infty$ corresponds to the uniform sampling.
The hyperparameter $\beta$ in Equation \eqref{eq_beta} controls how conservative the policy update is used during training.
If $\beta \rightarrow 0$, there are no constraints on policy updates, while $\beta \rightarrow \infty$ results in that agent learning is equivalent to behavior cloning.
The hyperparameter $\eta$ in Equation \eqref{eq_eta} and Equation \eqref{eq_beta} controls the importance weight of the uncertainty on actor-critic learning.
In all experiments, we choose $\alpha=0.2$, $\beta=0.1$ and $\eta=1$ according to Figure \ref{hyper}.

\ \newline
\noindent\textbf{E.3 Training curves}

\begin{figure}[H]
	\begin{center}
		\includegraphics[width=0.815\linewidth]{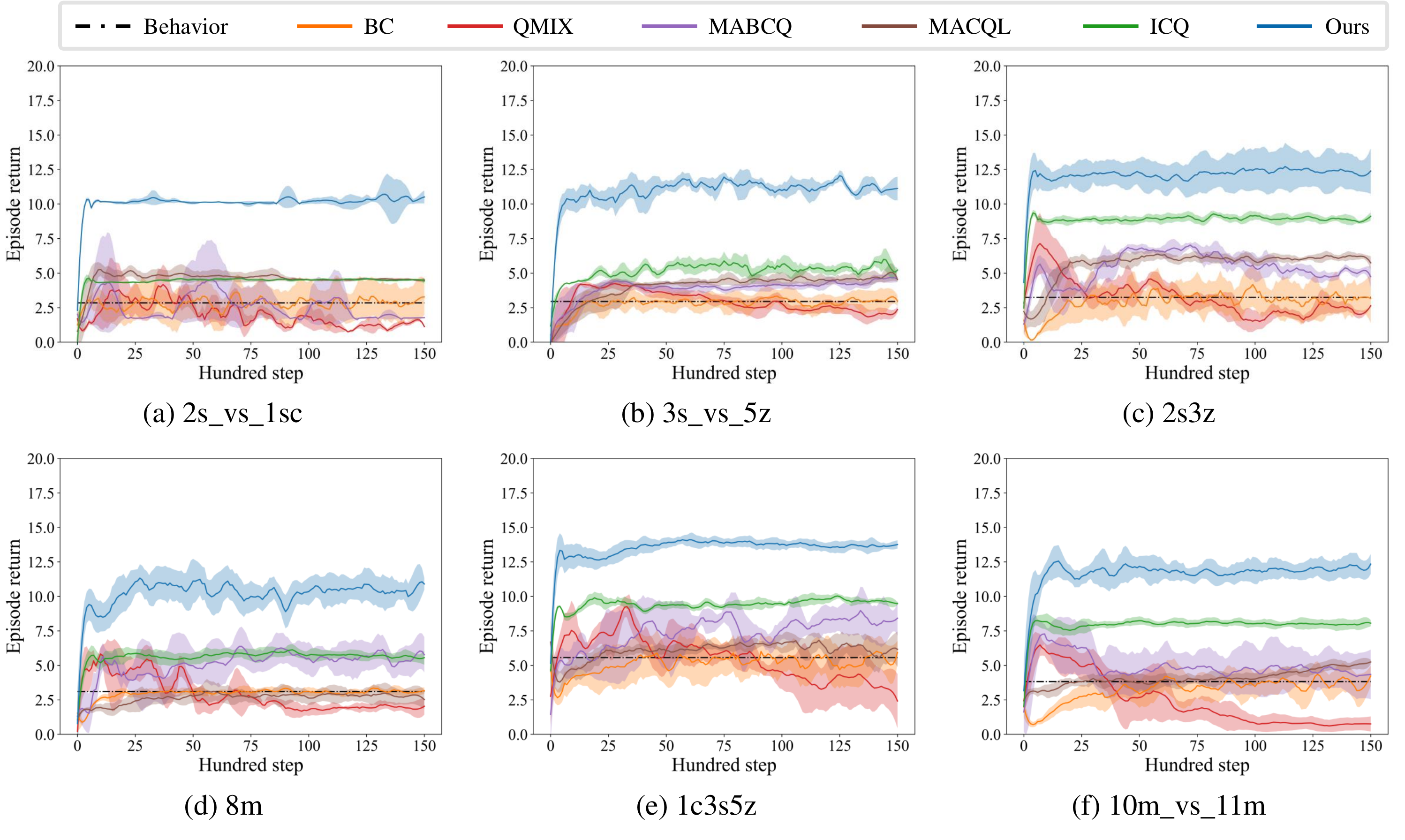}
		\caption{The performance of different algorithms on the low-quality datasets (StarCraft II)}
	\end{center}
\end{figure}

\begin{figure}[H]
	\begin{center}
		\includegraphics[width=0.815\linewidth]{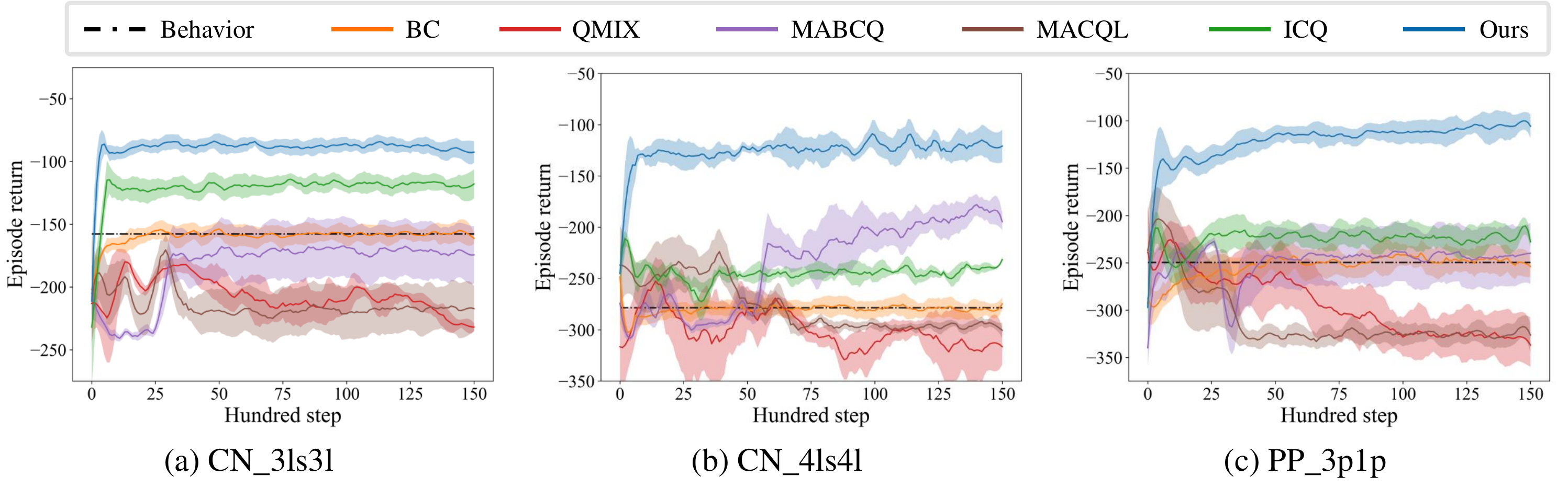}
		\caption{The performance of different algorithms on the low-quality datasets (MPE)}
	\end{center}
\end{figure}

\begin{figure}[H]
	\begin{center}
		\includegraphics[width=0.55\linewidth]{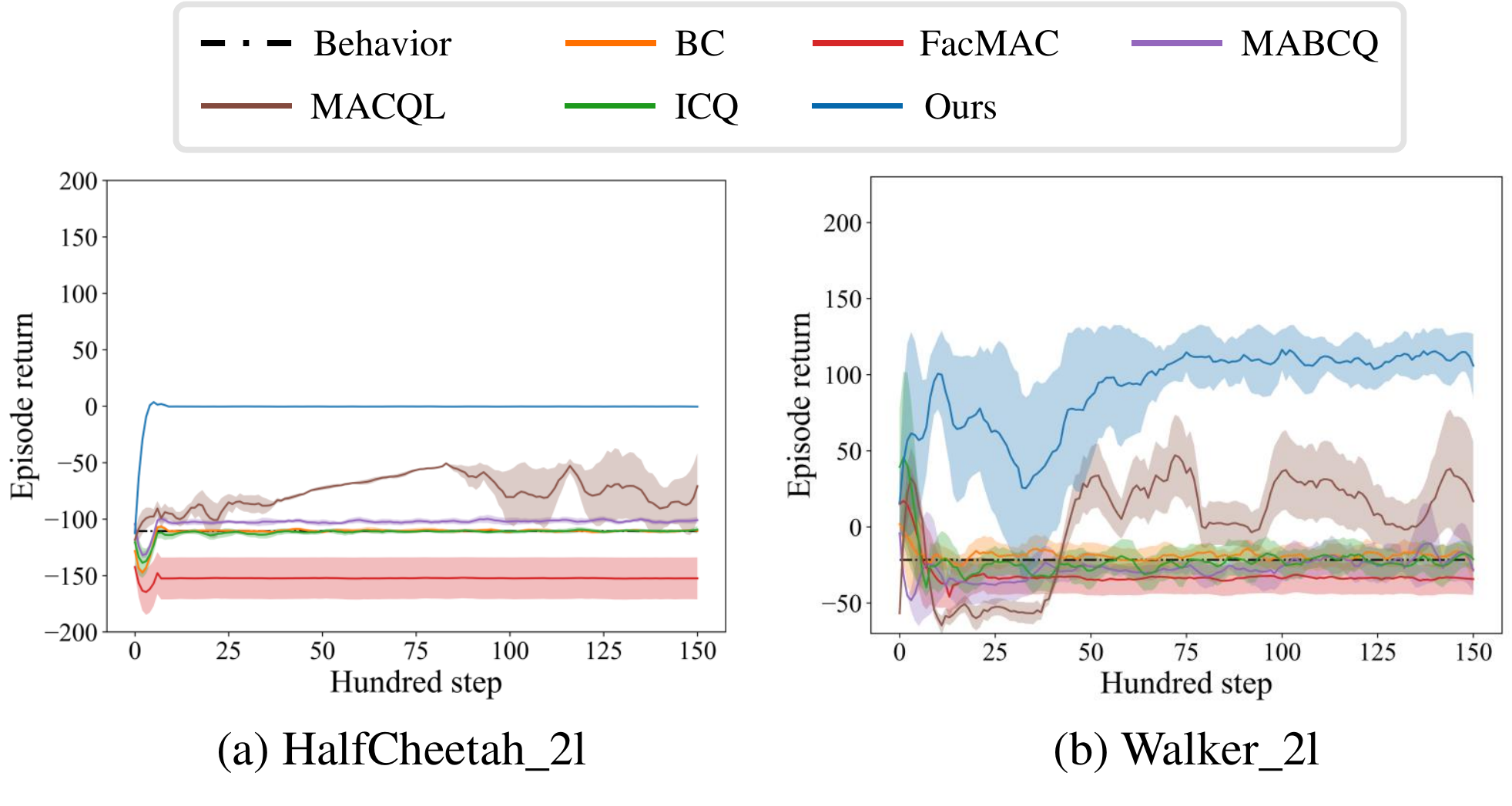}
		\caption{The performance of different algorithms on the low-quality datasets (MAmujoco)}
	\end{center}
\end{figure}

\begin{figure}[H]
	\begin{center}
		\includegraphics[width=0.815\linewidth]{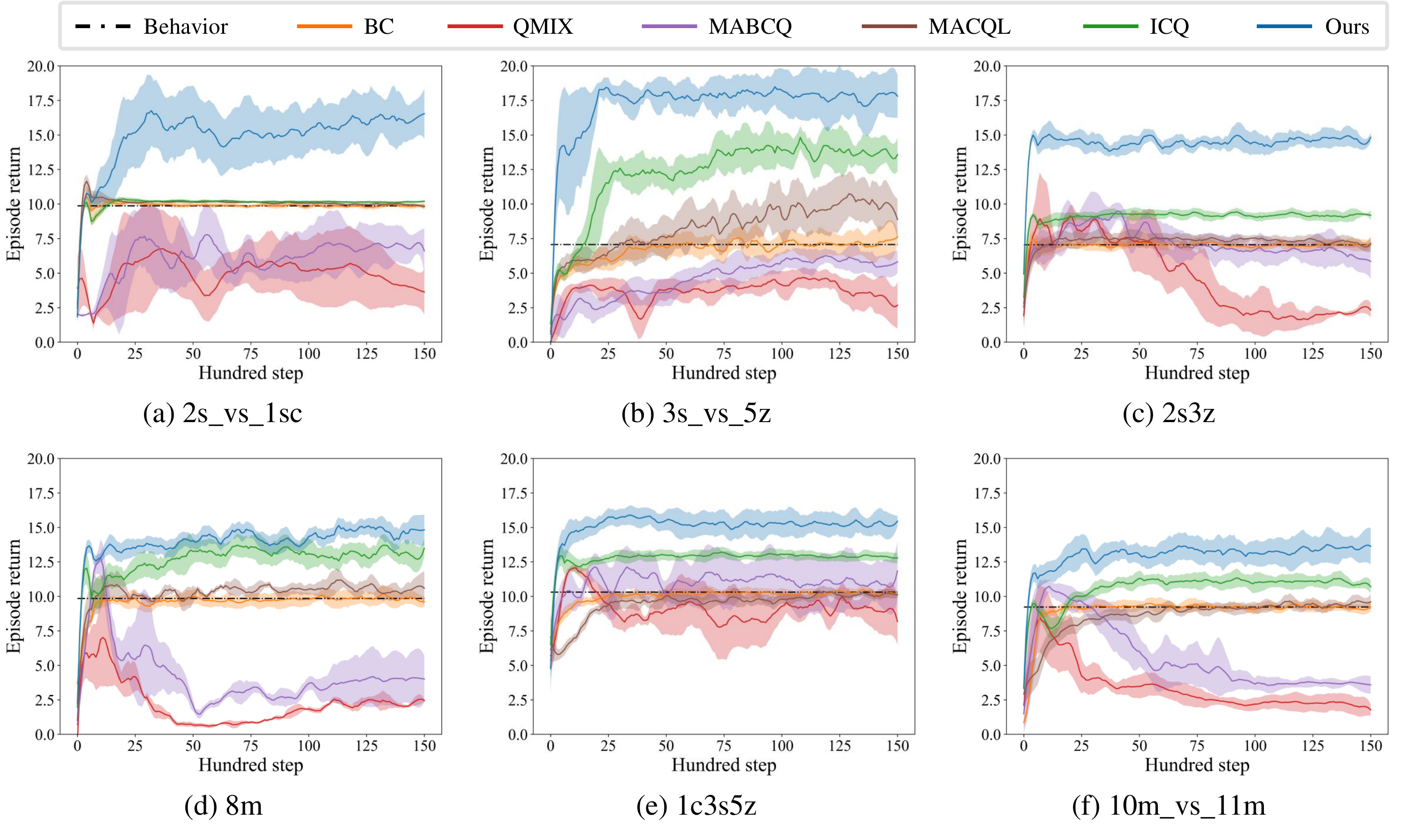}
		\caption{The performance of different algorithms on the medium-quality datasets (StarCraft II)}
	\end{center}
\end{figure}

\begin{figure}[H]
	\begin{center}
		\includegraphics[width=0.815\linewidth]{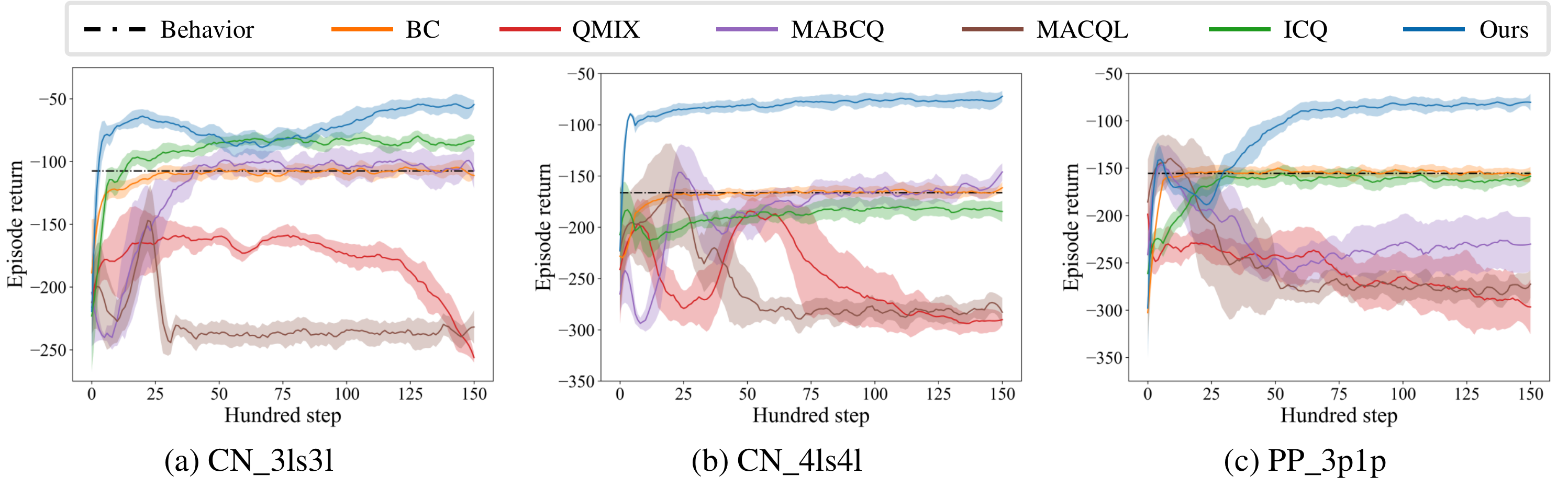}
		\caption{The performance of different algorithms on the medium-quality datasets (MPE)}
	\end{center}
\end{figure}

\begin{figure}[h]
	\begin{center}
		\includegraphics[width=0.55\linewidth]{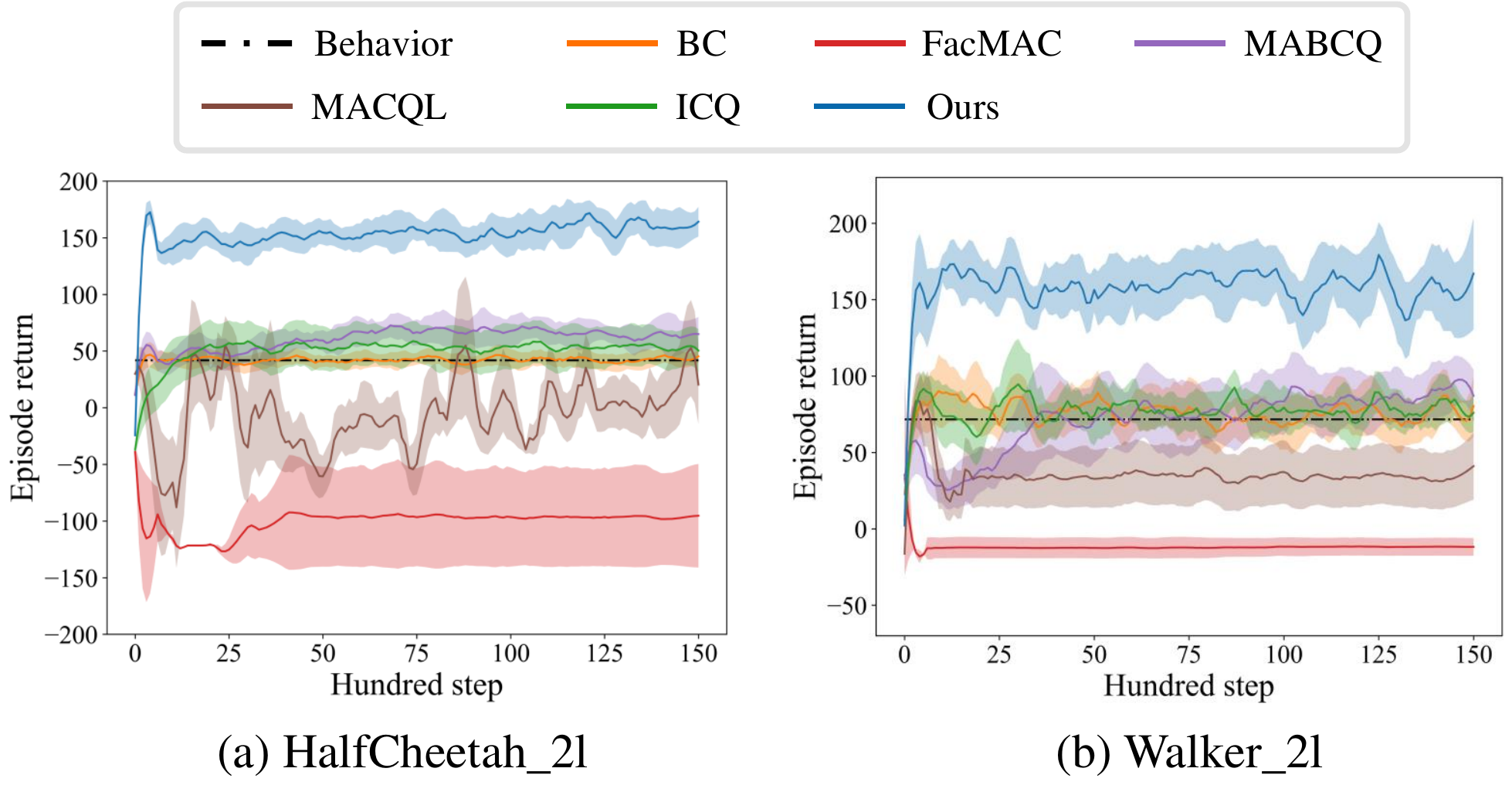}
		\caption{The performance of different algorithms on the medium-quality datasets (MAmujoco)}
	\end{center}
\end{figure}

\noindent\textbf{E.4 Decomposed weighted individual reward}

\begin{figure}[H]
	\begin{center}
		\includegraphics[width=0.815\linewidth]{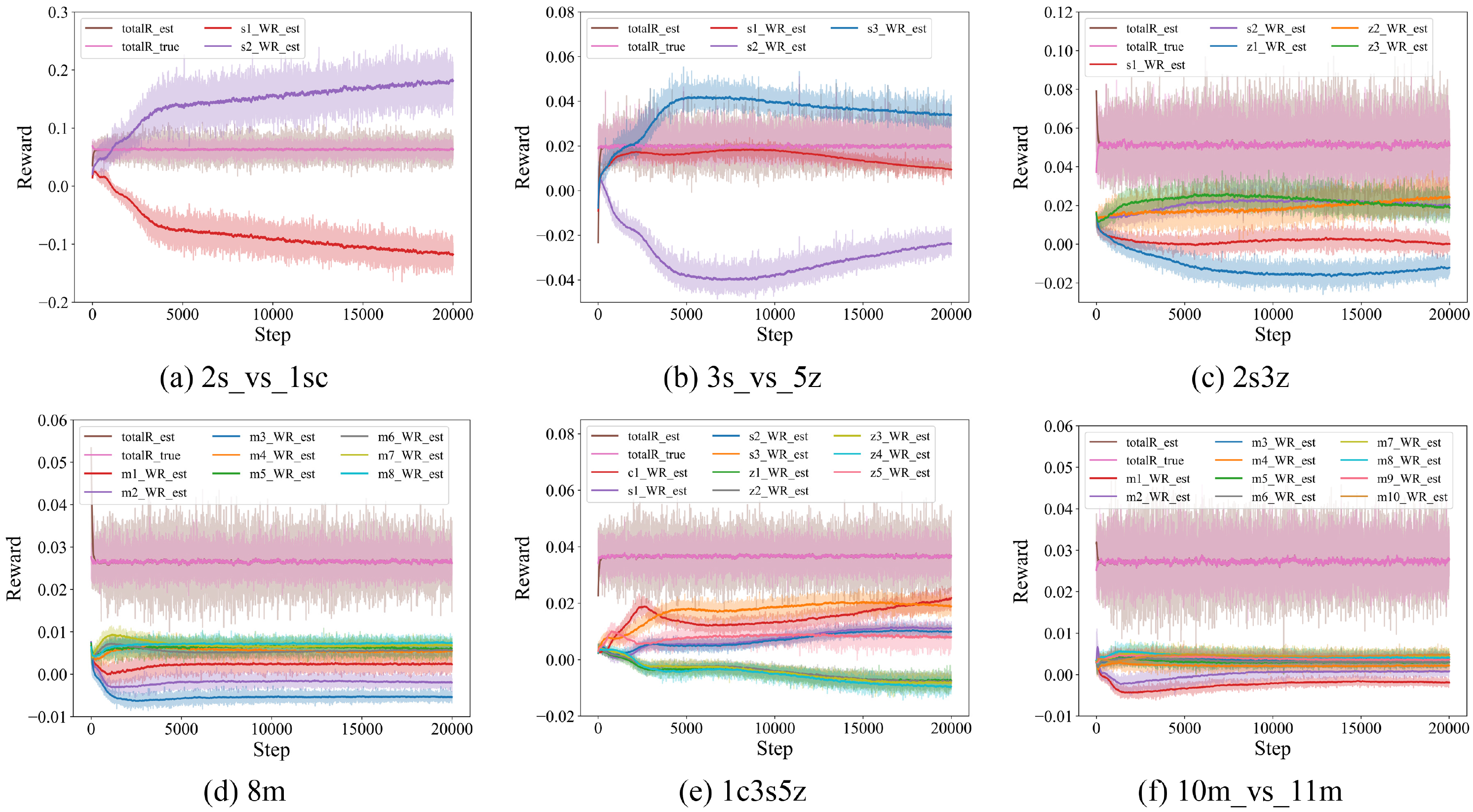}
		\caption{Decomposed weighted individual reward on the low-quality datasets (StarCraft II)}
	\end{center}
\end{figure}

\begin{figure}[H]
	\begin{center}
		\includegraphics[width=0.815\linewidth]{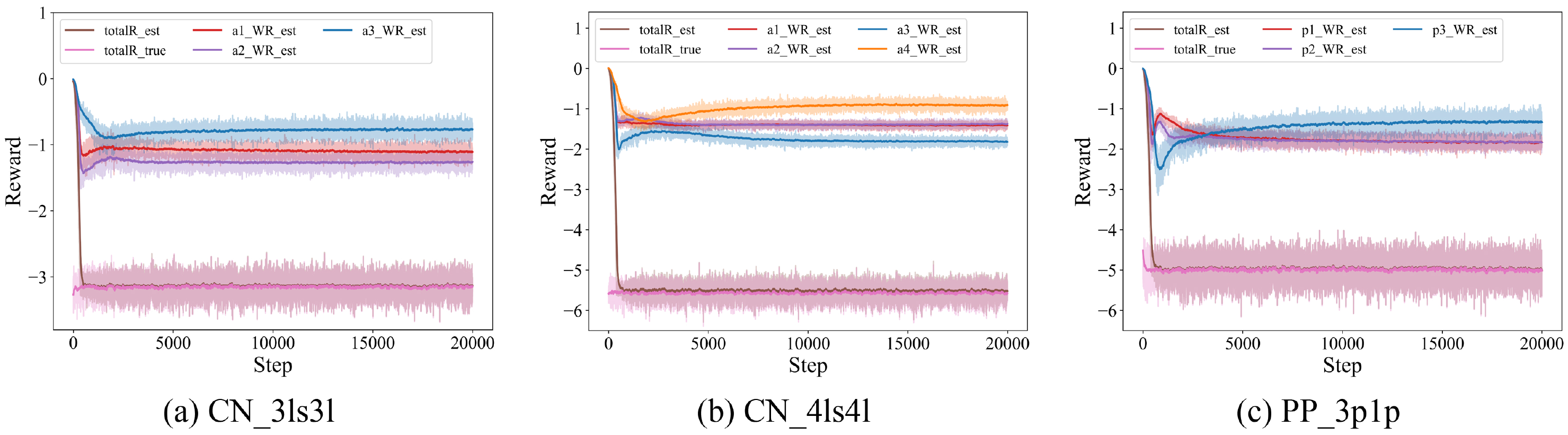}
		\caption{Decomposed weighted individual reward on the low-quality datasets (MPE)}
	\end{center}
\end{figure}

\begin{figure}[H]
	\begin{center}
		\includegraphics[width=0.55\linewidth]{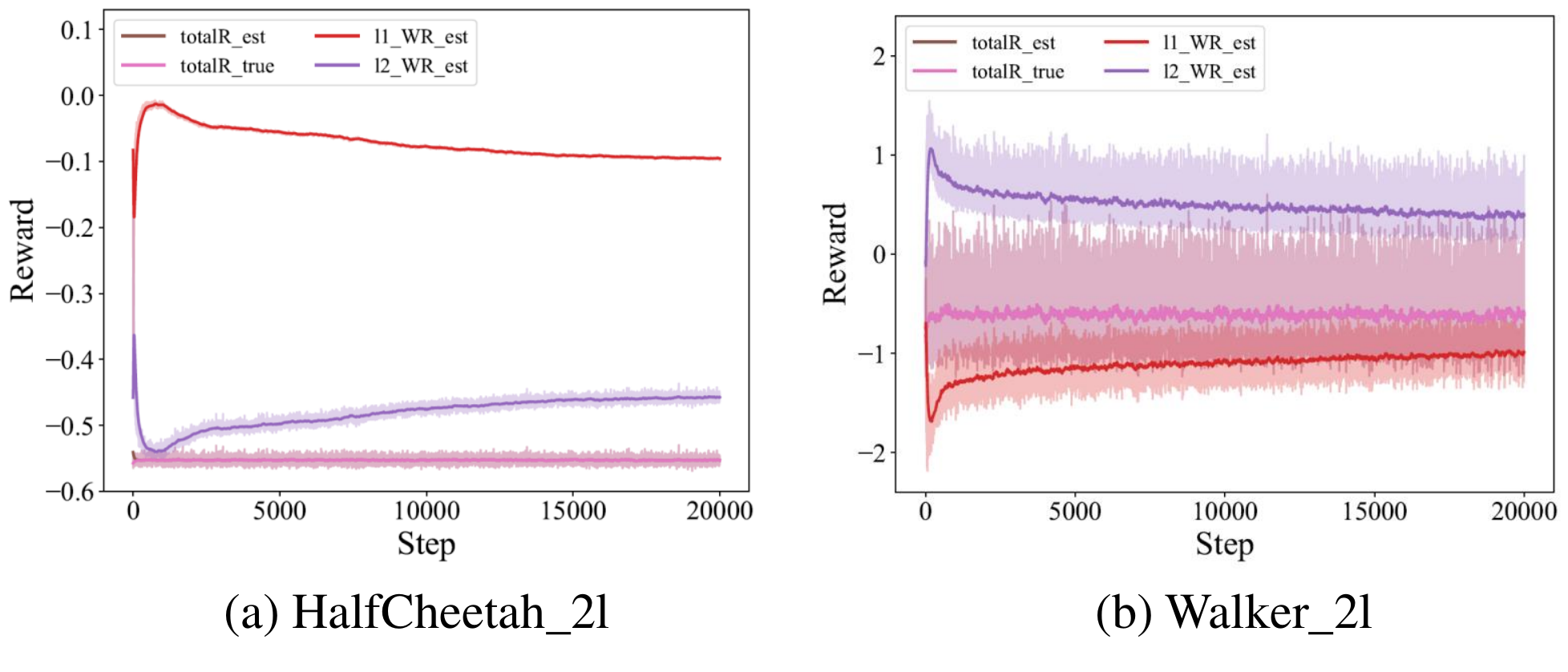}
		\caption{Decomposed weighted individual reward on the low-quality datasets (MAmujoco)}
	\end{center}
\end{figure}

\begin{figure}[h]
	\begin{center}
		\includegraphics[width=0.815\linewidth]{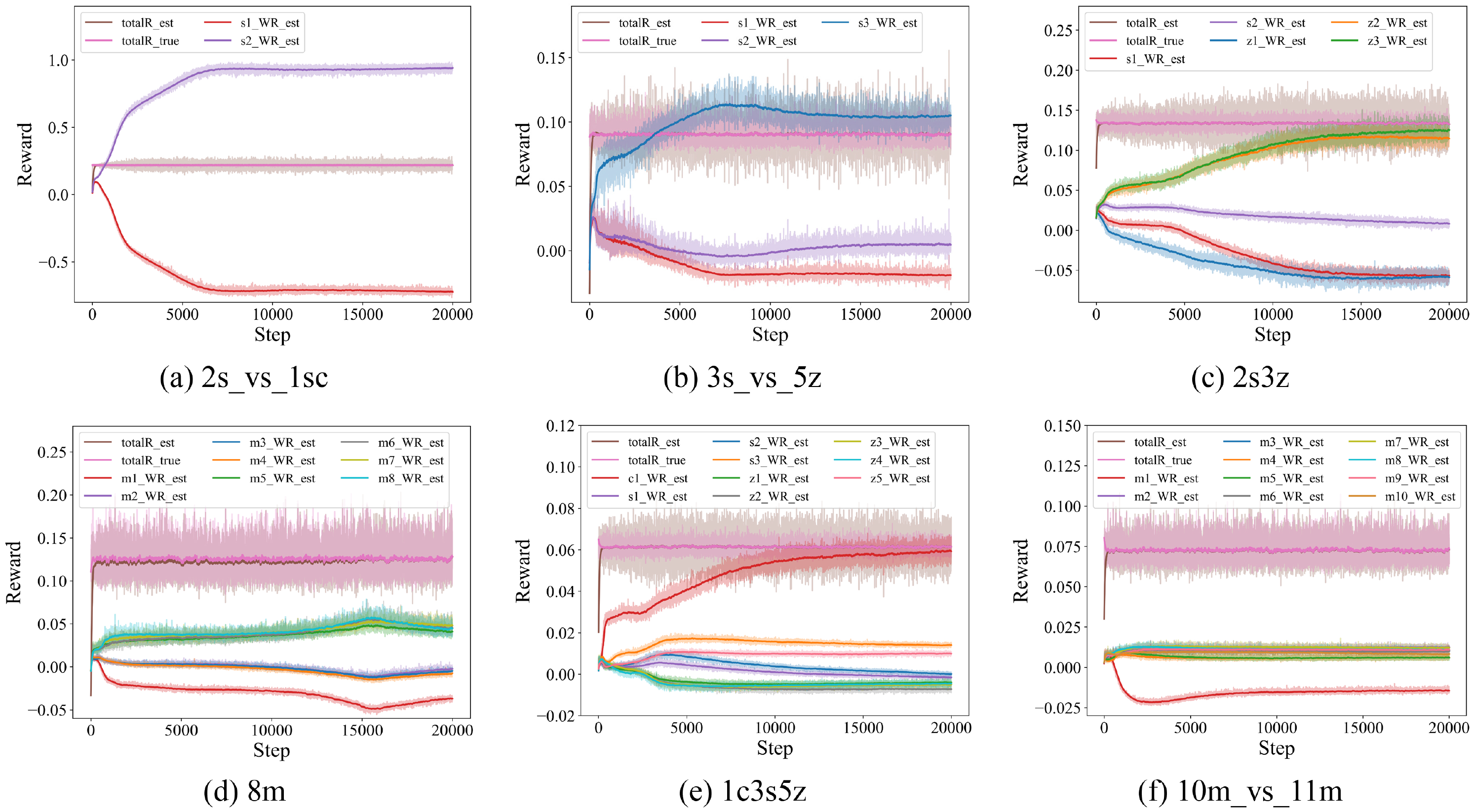}
		\caption{Decomposed weighted individual reward on the medium-quality datasets (StarCraft II)}
	\end{center}
\end{figure}

\begin{figure}[H]
	\begin{center}
		\includegraphics[width=0.815\linewidth]{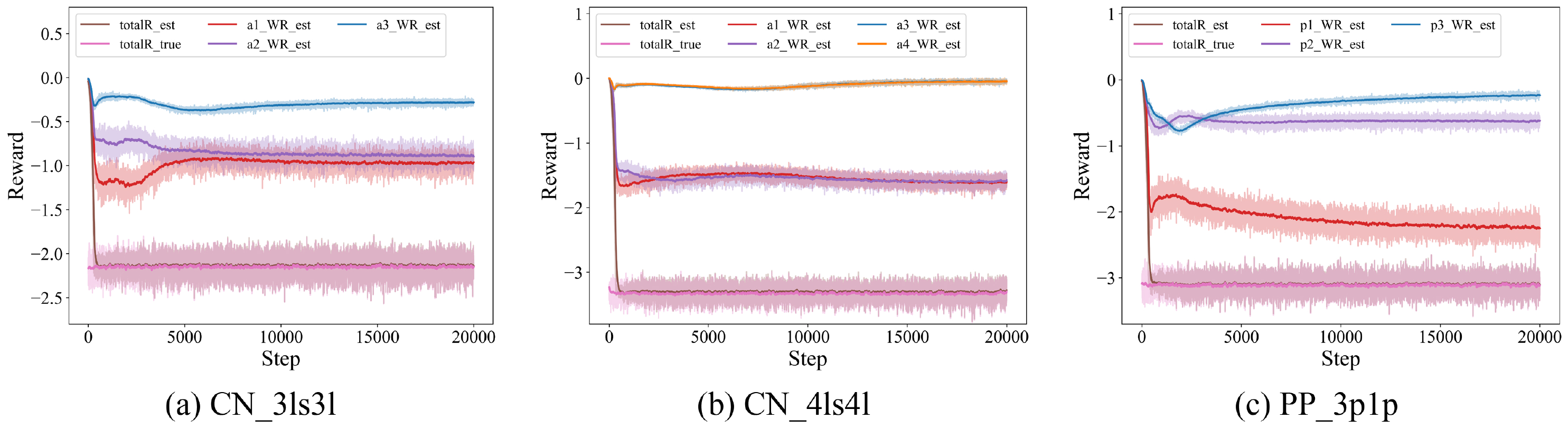}
		\caption{Decomposed weighted individual reward on the medium-quality datasets (MPE)}
	\end{center}
\end{figure}

\begin{figure}[H]
	\begin{center}
		\includegraphics[width=0.55\linewidth]{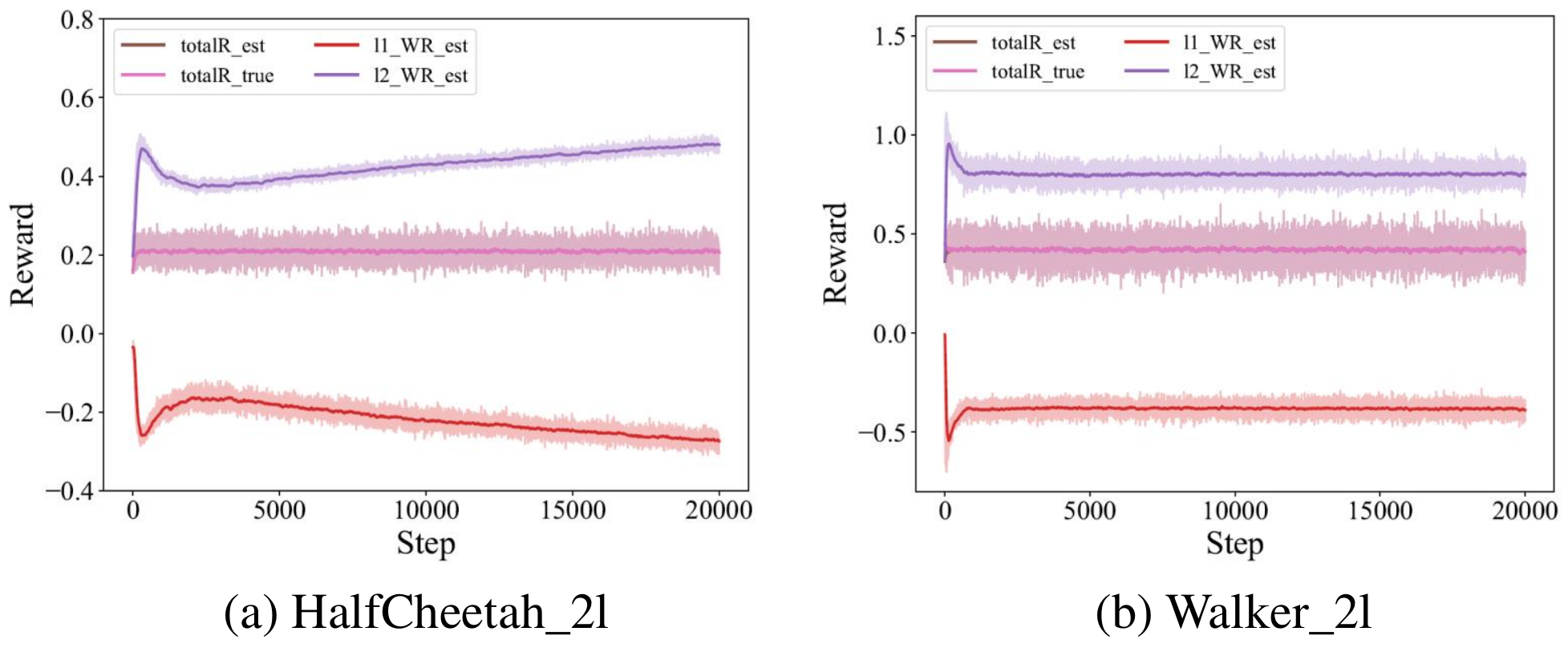}
		\caption{Decomposed weighted individual reward on the medium-quality datasets (MAmujoco)}
	\end{center}
\end{figure}

\end{document}